\documentclass[11pt]{article}

\usepackage[final]{acl}
\usepackage{times}
\usepackage{latexsym}

\usepackage[T1]{fontenc}

\usepackage[utf8]{inputenc}

\usepackage{microtype}

\usepackage{inconsolata}

\usepackage{graphicx}
\usepackage{booktabs}
\usepackage{amsmath}
\usepackage[table]{xcolor} 
\usepackage{colortbl}      
\usepackage{multirow}
\usepackage{amsfonts} 
\usepackage[most]{tcolorbox}
\usepackage{tcolorbox}
\usepackage{enumitem}
\usepackage{listings}
\usepackage{graphicx}
\usepackage{subcaption}

\definecolor{jsonblue}{RGB}{223,235,247}
\definecolor{LakeBlue}{RGB}{0,61,153}

\usetikzlibrary{shadows}

\lstdefinestyle{json}{
  basicstyle=\ttfamily\small,
  breaklines=true,
  frame=none,
  aboveskip=0pt,
  belowskip=0pt,
  showstringspaces=false
}

\title{Unified Thinker: A General Reasoning Modular Core for Image Generation}

\author{
  \textbf{Sashuai Zhou\textsuperscript{1,2,*}},
  \textbf{Qiang Zhou\textsuperscript{2,*}},
  \textbf{Jijin Hu\textsuperscript{2,*}},
  \textbf{Hanqing Yang\textsuperscript{2,*}},
  \textbf{Yue Cao\textsuperscript{3}},
  \textbf{Junpeng Ma\textsuperscript{4}}\\
  \textbf{Yinchao Ma\textsuperscript{2}},
  \textbf{Jun Song\textsuperscript{2,$\dagger$}},
  \textbf{Tiezheng Ge\textsuperscript{2}},
  \textbf{Cheng Yu\textsuperscript{2}},
  \textbf{Bo Zheng\textsuperscript{2}},
  \textbf{Zhou Zhao\textsuperscript{1,$\dagger$}}\\[2pt]
  \textsuperscript{1}Zhejiang University,
  \textsuperscript{2}Alibaba Group,
  \textsuperscript{3}Nanjing University,
  \textsuperscript{4}Fudan University\\
  \small{$^*$Equal contribution. \quad $^\dagger$Corresponding authors.}\\
  \small{\textbf{Project:} \url{https://github.com/LivingFutureLab/UnifiedThinker}}
}

\begin{document}
\maketitle
\begin{abstract}

Despite impressive progress in high-fidelity image synthesis, generative models still struggle with logic-intensive instruction following, exposing a persistent reasoning--execution gap. Meanwhile, closed-source systems (e.g., Nano Banana) have demonstrated strong reasoning-driven image generation, highlighting a substantial gap to current open-source models. We argue that closing this gap requires not merely better visual generators, but executable reasoning: decomposing high-level intents into grounded, verifiable plans that directly steer the generative process. To this end, we propose \textbf{Unified Thinker}, a task-agnostic reasoning architecture for general image generation, designed as a unified planning core that can plug into diverse generators and workflows. Unified Thinker decouples a dedicated Thinker from the image Generator, enabling modular upgrades of reasoning without retraining the entire generative model. We further introduce a two-stage training paradigm: we first build a structured planning interface for the Thinker, then apply reinforcement learning to ground its policy in pixel-level feedback, encouraging plans that optimize visual correctness over textual plausibility. Extensive experiments on text-to-image generation and image editing show that Unified Thinker substantially improves image reasoning and generation quality.


\end{abstract}
\begin{figure}[t]
  \includegraphics[width=\columnwidth]{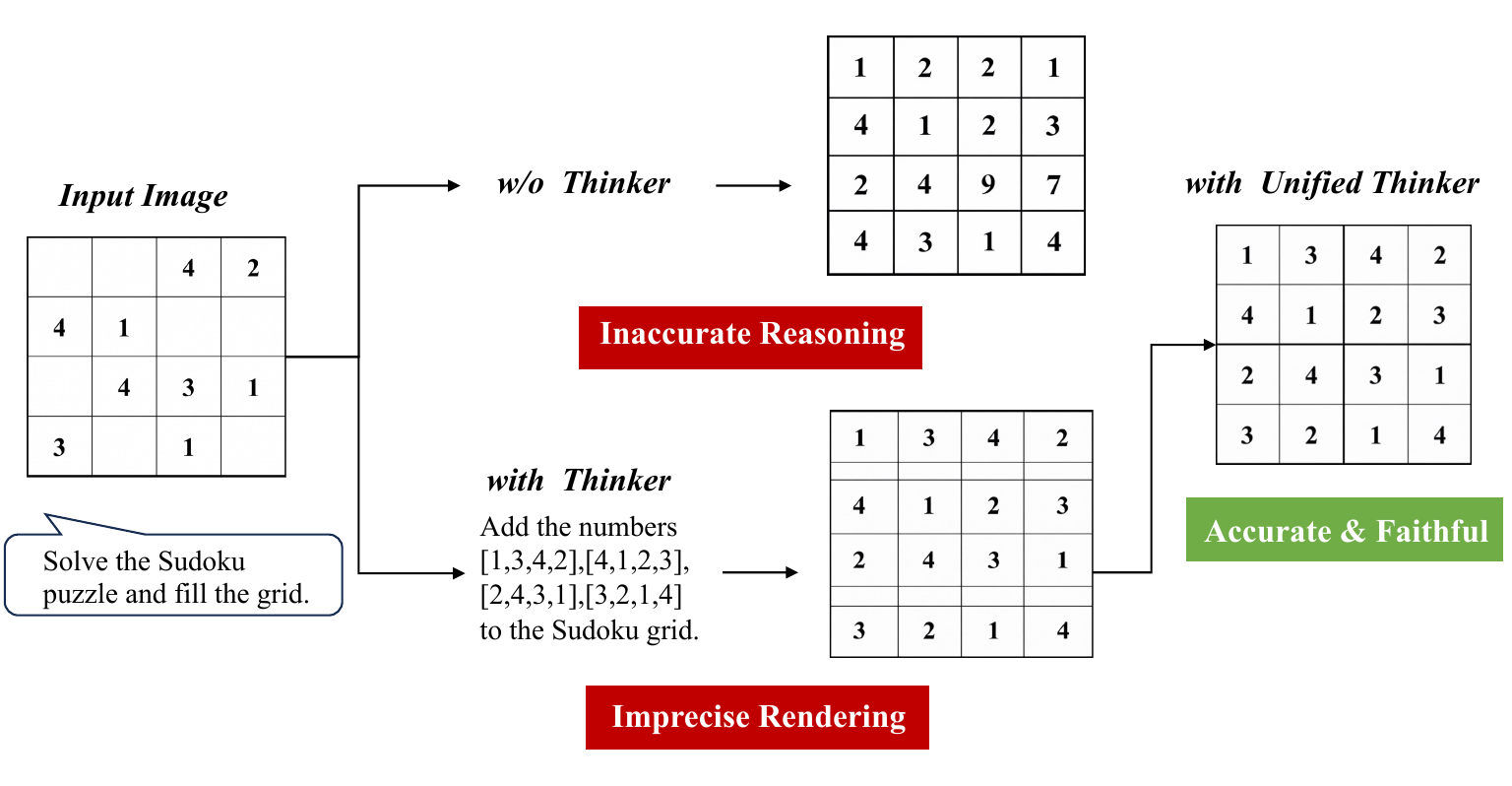}
  \caption{Challenges in reasoning-aware image generation. Existing models, exemplified by Qwen-Image-Edit, exhibit two failure modes: (1) inaccurate reasoning (without Thinker), leading to logically incorrect edits; and (2) imprecise rendering (with Thinker), where correct reasoning does not translate into faithful visual outputs. Our Unified Thinker aims to address both issues.}
  \label{fig:introduction}
\end{figure}

\begin{figure*}[t]
  \includegraphics[width=\linewidth]{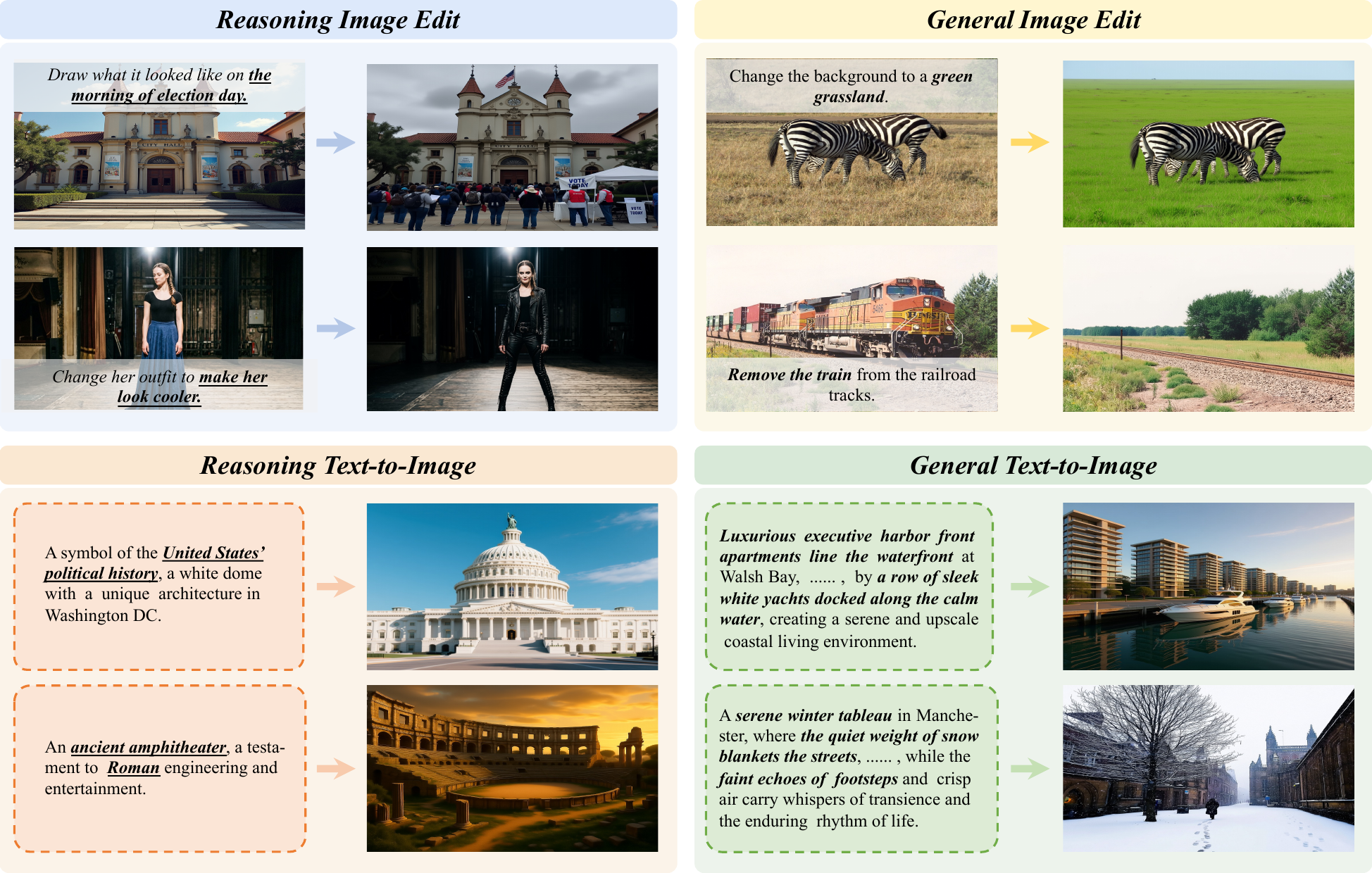}
  \caption{Visual demonstrations of Unified Thinker on unified image generative tasks, including image editing and text-to-image generation, along with reasoning.}
  \label{fig:intro}
\end{figure*}

\section{Introduction}


The rapid evolution of diffusion-based foundation models~\cite{ho2020denoising,dhariwal2021diffusion,rombach2022high,xu2024diffusion_dpo} has driven an unprecedented leap in high-fidelity image synthesis. Advanced proprietary models such as GPT-4o~\cite{hurst2024gpt4o} and Nano Banana~\cite{nanobanana} have recently demonstrated strong reasoning-driven image generation under complex instructions. In contrast, despite steady progress in open-source systems~\cite{sd3.5,blackforest2024flux,bagel2025,qwenimage-edit,liao2025imagegen-cot}, current open-source models still exhibit a clear gap in handling logic-intensive or implicit directives~\cite{wisebench,risebench,reason2edit,liu2025step1x-edit}.


Current attempts to bridge this gap follow two primary approaches. \textbf{Built-in Reasoning} internalizes reasoning into the generator via unified training that couples multimodal understanding with generation~\cite{bagel2025,show-o2,xiao2024omnigen}. However, this tight entanglement reduces modularity and may destabilize training, often degrading the generator’s visual fidelity. In contrast, \textbf{External Planner-Driven} methods use an MLLM to plan for a mostly frozen generator~\cite{qwenimage-edit,lin2025uniworld,li2025editthinker,reason2edit}. While modular, they suffer from a reasoning--execution mismatch: text-space plans are not grounded in the generator’s capabilities, so even correct plans can cause visual failures, and iterative planning further increases compute.


We identify the key bottleneck as the absence of a principled paradigm for reasoning in image generation. In this paper, we propose \textbf{Unified Thinker}, a universal reasoning core for general image generation, built around a think-then-execute architecture that parametrically decouples the Thinker for instruction understanding and planning from the Generator for pixel synthesis. Here, the Generator refers to the underlying image synthesis backbone (e.g., a diffusion model) that takes conditioning signals and produces the final image in pixel space. The Thinker is implemented as a standalone, trainable multimodal large language model (MLLM) that transforms an instruction into a hierarchical, generator-friendly plan consisting of an intent summary, explicit constraints, and ordered sub-goals, which the Generator consumes as conditioning, enabling strong task transferability across text-to-image and editing and plug-and-play compatibility with different generator backbones.

However, this decoupled design still faces additional challenges: as shown in Fig.~\ref{fig:introduction}, without proper alignment, a naive Thinker may produce plausible reasoning that the Generator cannot execute. To bridge the reasoning-to-execution gap, we introduce a dedicated data-to-training pipeline to align planning with visual outcomes. We first construct \textbf{HieraReason-40K}, a hierarchical reasoning dataset synthesized with Gemini-3-Pro~\cite{nanobanana}, which pairs complex instructions with structured, executable plans to teach the Thinker the desired planning format and basic logical decomposition. We then adopt a two-stage training strategy: we perform joint supervised fine-tuning on HieraReason-40K to establish initial plan quality, followed by an end-to-end dual-phase reinforcement learning procedure that places the Generator in the loop and optimizes the Thinker using rewards computed from the final image’s constraint satisfaction. This directly grounds the Thinker’s policy in pixel-level feedback, encouraging plans that are not only semantically plausible but also executable under the Generator’s capabilities.

We conduct extensive evaluations in four settings: text-to-image reasoning, reasoning-based image editing, general text-to-image generation, and general image editing. Across all benchmarks, Unified Thinker delivers substantial gains in generative reasoning, markedly improving instruction following and constraint satisfaction. These improvements also hold across multiple generator backbones, supporting our core claim that a decoupled Thinker learns reusable, executable reasoning patterns that transfer across models and tasks.

Our main contributions are as follows:

\begin{itemize}
    \item We propose a decoupled reasoning-generation framework \textbf{Unified Thinker} that utilizes a unified module to handle general image generation tasks, significantly enhancing modular adaptability and transferability.
    \item We introduce an end-to-end training pipeline spanning from hierarchical reason data construction to execution-led reinforcement learning, bridging the gap between abstract reasoning and pixel-level execution.
    \item Through comprehensive experimental results, we demonstrated a significant performance improvement in reasoning-intensive generation tasks and verified the cross-model portability of our reasoning core module.
\end{itemize}
\section{Related Work}

\subsection{Foundational Generative Models}
Modern image generation is predominantly anchored in diffusion-based frameworks~\cite{ho2020denoising,rombach2022high}. Recent advances~\cite{sd3.5,blackforest2024flux,qwenimage-edit} build upon Diffusion Transformers~\cite{peebles2023scalable} and flow matching~\cite{flowmatch} to improve fidelity, prompt alignment, and diversity in latent diffusion models.  Meanwhile, an emerging direction unifies autoregressive modeling with visual generation in a single framework, giving rise to unified multimodal models~\cite{bagel2025,janus,show-o2,xiao2024omnigen}. For instance, Bagel~\cite{bagel2025} uses a transformer backbone to jointly model text and image tokens, whereas OmniGen~\cite{xiao2024omnigen} dispenses with external encoders and handles multiple vision tasks through a unified pipeline. In parallel, image editing has evolved from mask-based inpainting~\cite{zhuang2024task,ju2024brushnet} to instruction-guided manipulation~\cite{brooks2023instructpix2pix,yu2025anyedit}. To further enhance instruction following, recent methods~\cite{li2024smartedit, fu2024guiding, lin2025uniworld,liu2025step1x-edit} such as Qwen-Image-Edit~\cite{qwenimage-edit} leverage MLLMs for instruction parsing and planning. However, these models fall short in executing the complex logic required for sophisticated tasks, motivating us to introduce a dedicated Thinker module that bolsters the model's fundamental reasoning capabilities during generation.


\subsection{Reasoning for Image Generation}
Recent research has moved beyond the one-shot mapping paradigm by explicitly incorporating reasoning into the image generation process. One line of work~\cite{t2ir1,wang2025mint,liao2025imagegen-cot,zhou2026spatialreward,unicot} introduces clear intermediate representations to decompose complex prompts into structured steps or explicit spatial layouts, improving compositional consistency and coherence.
Another line of work~\cite{reason2edit,mi2025thinkdiff,bagel2025} encourages models to reason about intent and constraints before drawing, moving beyond one-shot planning to better satisfy complex requirements, like R-Genie~\cite{rgenie}, which infers latent user intent instead of merely following the surface-level prompt.
A third line of work~\cite{guo2025canwegen,wu2025omnigen2,li2025reflectdit,li2025editthinker,yin2025reasonedit} focuses on post-generation refinement by introducing reflection-and-correction mechanisms that assess the generated image, diagnose issues, and iteratively update the output to improve final quality. For example, Reflect-DiT~\cite{li2025reflectdit} introduces explicit self-reflection to guide revision, while EditThinker~\cite{li2025editthinker} enables reasoning via multi-round reflective interactions throughout the editing process. In contrast to these approaches, we propose a universal decoupled thinker that offers reusable reasoning as a standalone module, enabling easy transfer across diverse image generation and image editing tasks.

\section{Data Construction}
\label{sec:data}

\paragraph{Goal and dataset.}
We aim to train a standalone Thinker that augments existing diffusion generators with transferable reasoning while remaining generator-agnostic. To this end, we construct \textbf{HieraReason-40K}, a selected general-purpose corpus by combining four sources that cover text-to-image generation, general image editing, reasoning image generation, and reasoning image editing tasks~\cite{han2025unireditbench,huang2025interleaving,qian2025pico,fang2025fluxreason}. Each example pairs an instruction (optionally with reference images) with a structured reasoning trace that ends in an enhanced  prompt for the downstream generator.

\begin{figure}[t]
  \includegraphics[width=\columnwidth]{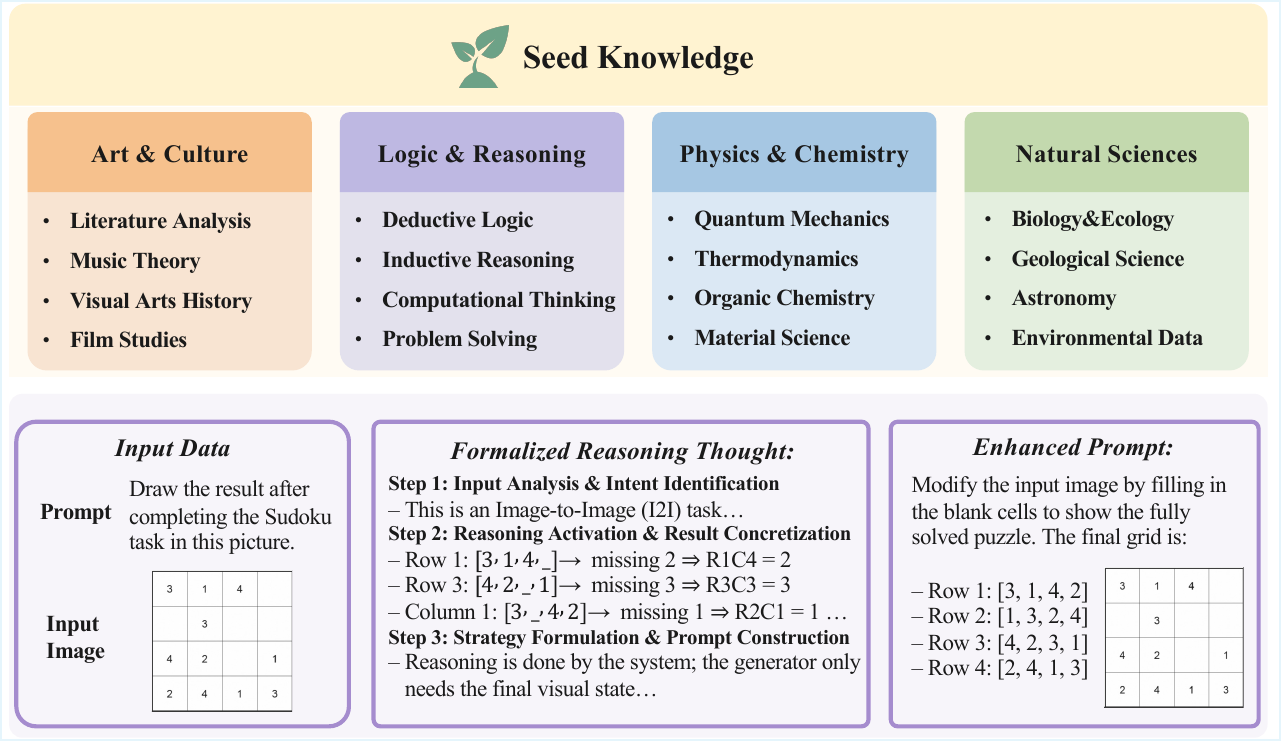}
  \caption{Data construction pipeline for HieraReason-40K. We combine seed knowledge and user requests to generate structured reasoning traces and executable enhanced prompts.}
  \label{fig:data}
\end{figure}

\paragraph{Structured reasoning trace.}
As illustrated in Fig.~\ref{fig:data}, we create inference-style supervision by combining broad seed knowledge (e.g., art \& culture) with input instruction to form generated inference data. Each training example is then rewritten into a rigorous structured reasoning trace: given an original instruction (and an optional reference image for image editing), the annotator produces a formalized reasoning trace followed by a final enhanced prompt for the generator. The trace follows a fixed three-stage procedure. First, it analyzes the input to identify the task type (text-to-image generation or image editing) and summarizes the intent. 
Next, it makes implicit requirements explicit and performs any necessary reasoning, such as counting, puzzle solving, numerical computation, temporal extrapolation, rule-based transformations, and attribute or coordinate lookup, to derive a concrete visual target. Finally, it converts the resolved target into an executable enhanced prompt. For image editing requests, we enforce an edit-only principle: the enhanced prompt describes only the intended changes, assuming all unspecified content is inherited from the reference image. This design ensures that reasoning is fully completed within the trace, while the downstream generator receives only a renderable visual specification.

\paragraph{Annotation and quality control.}
We use Gemini3-Pro~\cite{nanobanana} to generate initial structured reasoning traces, followed by automatic normalization to enforce strict format consistency (e.g., mandatory stage headers and standardized image placeholders such as \texttt{<image>}). We further filter or rewrite samples that violate the trace format, fail to follow the edit-only principle for image editing, produce non-visual or underspecified targets, or exhibit inconsistencies between the reasoning trace and the final enhanced prompt. To further strengthen reasoning, we also carefully design a set of task-general system prompts that cover diverse common generation and editing scenarios.

\section{Framework and Training}
\label{sec:method}

The core objective of our framework is to mitigate the reasoning–execution mismatch in reasoning-driven image generation and editing. We introduce a decoupled think-then-execute framework with two components: \textbf{Thinker}, a standalone, trainable multimodal large language model that produces structured reasoning traces and an executable visual specification, and \textbf{Generator}, a diffusion-based model that synthesizes the final image conditioned on the Thinker’s outputs. Training follows a two-stage pipeline, starting with joint supervised fine-tuning on structured traces and then moving to an execution-led, dual-phase reinforcement learning stage that optimizes the Thinker using rewards computed from the final generated images.




\subsection{Joint Supervised Fine-Tuning}
\label{subsec:sft}

To teach the Thinker a consistent reasoning format and establish the think-then-execute pipeline, we first perform joint supervised fine-tuning stage. Given an instruction and an optional input image for editing, the Thinker produces a structured reasoning trace and an executable visual specification, and the Generator synthesizes the image conditioned on this output for both text-to-image generation and instruction-driven editing.

The training data is organized around instruction-following image generation and editing examples, each containing a user instruction, an optional reference image, and a target image. We derive two synchronized views of the same examples for joint training: (1) an \textit{understanding view}, which pairs the input (instruction and optional reference image) with the annotated structured reasoning trace, supervising the Thinker via a language modeling loss; and (2) a \textit{generation view}, which pairs the executable enhanced prompt (extracted from the trace) with the target image, supervising the Generator via the standard diffusion denoising objective. During each training step, we sample mini-batches from the two views and optimize a weighted sum of the understanding loss $\mathcal{L}_{\text{und}}$ (token-level cross-entropy) and the generation loss $\mathcal{L}_{\text{gen}}$ (noise-prediction mean squared error).

\begin{figure*}[th]
  \includegraphics[width=\linewidth]{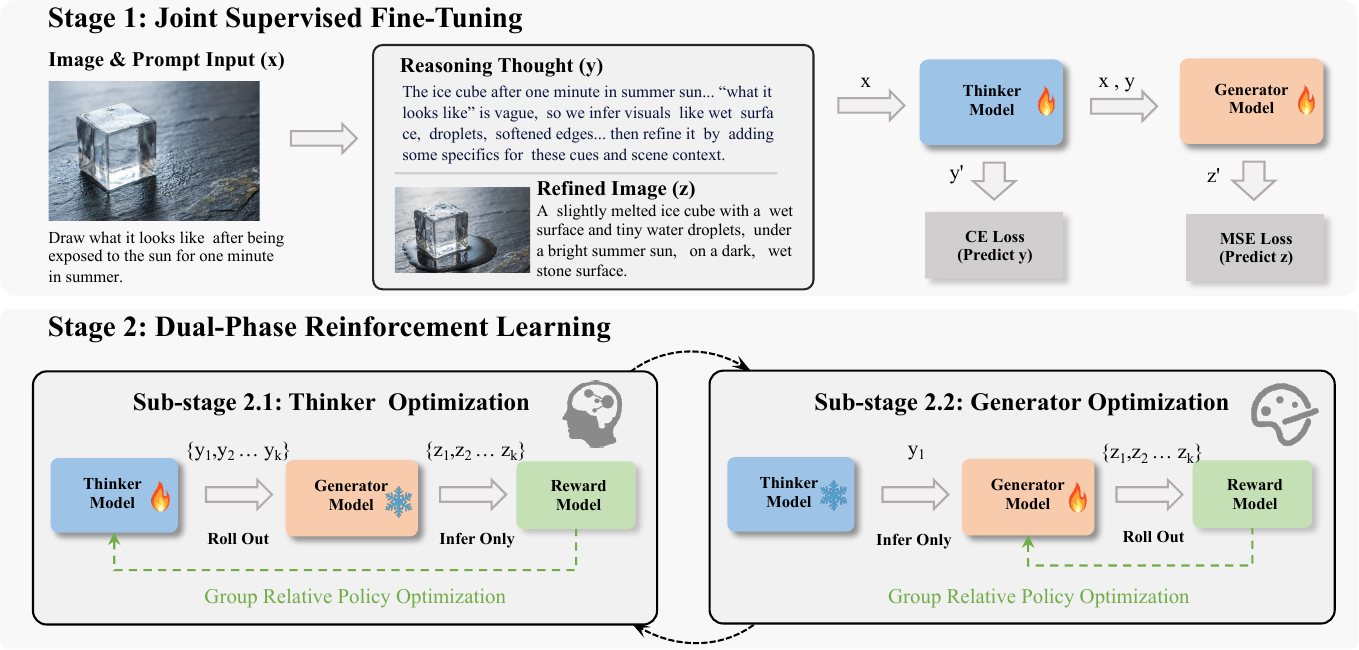}
  \caption {Our proposed two-stage framework for reasoning-aware image generation. Stage 1 initializes the Thinker Model and Generator Model. Given an Image \& Prompt (x), the Thinker generates a Reasoning Thought (y), which then guides the Generator to produce a Refined Image (z). Stage 2 further refines the Thinker and Generator Models to enhance their capability in integrating complex reasoning (y) into high-fidelity visual outputs (z), applicable to both novel image generation and existing image editing tasks.
}
\end{figure*}

This joint supervised fine-tuning procedure effectively aligns the instruction-generation capability of the Thinker with the image-synthesis prior of the Generator, ensuring that the produced reasoning instructions are not only semantically accurate but also highly compatible with the Generator’s operational semantics, thereby laying a solid foundation for cascaded inference deployment.

Formally, the overall objective is defined as:
\begin{equation}
\begin{aligned}
\mathcal{L}_{\text{SFT}} =\; &
\mathcal{L}_{\text{gen}}\big(\text{Generator}(\mathbf{y},\, \mathbf{x}_{\text{ref}}),\, \mathbf{x}_{\text{tgt}}\big) \\
& + \lambda \, \mathcal{L}_{\text{und}}\big(\text{Thinker}(\mathbf{x}_{\text{img}}),\, \mathbf{y}\big),
\end{aligned}
\end{equation}

where $\mathbf{x}_{\text{img}}$ denotes the input image, $\mathbf{y}$ is the ground-truth reasoning process, $\mathbf{x}_{\text{ref}}$ represents an optional reference image, $\mathbf{x}_{\text{tgt}}$ is the target output image, and $\lambda > 0$ is a hyperparameter balancing the two learning signals.

\subsection{Dual-Phase Reinforcement Learning}

While joint fine-tuning provides an initial alignment, it leaves a nontrivial reasoning--execution gap: the Thinker may produce plans that are plausible in text but suboptimal for the generator to execute. To address this without additional manual annotation, we introduce a dual-phase reinforcement learning strategy based on Group Relative Policy Optimization~\cite{guo2025deepseekr1}. The key idea is to sample multiple candidate traces for the same request, execute them with the generator, and train the Thinker by relative advantage feedback, promoting outputs that lead to better images and suppressing those that do not.

\textbf{Phase 1: Reasoning-Oriented RL.} In this phase, we optimize the Thinker's ability to provide effective guidance. For a given instruction, the Thinker samples a group of $G$ reasoning paths $\{y_1, y_2, \dots, y_G\}$. We use the Generator (fixed) to produce the corresponding images and assign a reward $r_i$ to each path based on the final image quality. We optimize the Thinker by maximizing:
\begin{equation}
\mathcal{J}_T(\theta_T) = \mathbb{E} \left[ \frac{1}{G} \sum_{i=1}^{G} \left( \frac{\pi_{\theta}(y_i|p)}{\pi_{old}(y_i|p)} \cdot \hat{A}_i \right) \right]
\end{equation}
\begin{equation}
\hat{A}_i = \frac{r_i - \text{mean}(\{r\})}{\text{std}(\{r\})}
\end{equation}

Here, $\hat{A}_i$ is the relative advantage, which tells the model which reasoning chains performed better than the group average. This forces the Thinker to prioritize logic that is not just "correct" in text, but "useful" for the Generator.

\textbf{Phase 2: Generation-Oriented RL.} With the Thinker providing reliable plans, we next improve the Generator’s execution fidelity. However, probability-flow ODE sampling in diffusion models is essentially deterministic, limiting the stochastic rollouts required by reinforcement learning. Following a Flow-GRPO-like idea~\cite{flowgrpo}, we convert the ODE sampler into an equivalent reverse-time SDE to introduce controlled randomness, enabling $G$ distinct rollouts $\{z_1, z_2, \dots, z_G\}$ for the same instruction and optimizing the Generator accordingly:
\begin{equation}
\mathcal{J}_G(\theta_G) = \mathbb{E} \left[ \frac{1}{G} \sum_{i=1}^{G} \left( \frac{\pi_{\theta}(z_i|c, p)}{\pi_{old}(z_i|c, p)} \cdot \hat{A}_i \right) \right]
\end{equation}
In this stage, the advantage $\hat{A}_i$ assigns higher credit to denoising trajectories that yield better images. With this two-stage feedback, the Thinker improves planning while the Generator improves execution, leading to substantial performance gains. Details of the reward design are provided in the appendix~\ref{sec:reward_model}.


\begin{table*}[thb]
\centering
\caption{Performance comparison of models on the RiseBench benchmark. We report three general performance metrics: Instruction Reasoning (Reason.), Appearance Consistency (Consist.), and Visual Plausibility (Visual.). Additionally, we present category-wise accuracy (\%) for four specific reasoning dimensions: Temporal, Causal, Spatial, and Logical. The \textit{Overall} score is the average of these four category-wise accuracies.}
\label{tab:risebench}
\fontsize{10pt}{12pt}\selectfont  
\setlength{\tabcolsep}{4pt}
\begin{tabular}{lccc|ccccc}
\toprule
\textbf{Model} & \textbf{Reason.} & \textbf{Consist.} & \textbf{Visual.} & \textbf{Temporal} & \textbf{Causal} & \textbf{Spatial} & \textbf{Logical} & \textbf{Overall} \\
\midrule
 Gemini-3-pro-image-preview & 77.0	&85.5	&94.4	&41.2	&61.1&	48.0	&37.6&	47.2 \\
Gemini-2.5-Flash-Image       & 61.2 & 86.0 & 91.3 & 25.9 & 47.8 & 37.0 & 18.8 & 32.8 \\
GPT-Image-1            & 62.8 & 80.2 & 94.9 & 34.1 & 32.2 & 37.0 & 10.6 & 28.9 \\
GPT-Image-1-mini       & 54.1 & 71.5 & 93.7 & 24.7 & 28.9 & 33.0 &  9.4 & 24.4 \\
Gemini-2.0-Flash-exp   & 48.9 & 68.2 & 82.7 &  8.2 & 15.5 & 23.0 &  4.7 & 13.3 \\
BAGEL (w/ CoT)         & 45.9 & 73.8 & 80.1 &  5.9 & 17.8 & 21.0 &  1.2 & 11.9 \\
Seedream-4.0           & 58.9 & 67.4 & 91.2 & 12.9 & 12.2 & 11.0 &  7.1 & 10.8 \\
Gemini-2.0-Flash-pre   & 49.9 & 68.4 & 84.9 & 10.6 & 13.3 & 11.0 &  2.3 &  9.4 \\
FLUX.1-Kontext-Dev     & 26.0 & 71.6 & 85.2 &  2.3 &  5.5 & 13.0 &  1.2 &  5.8 \\
Ovis-U1                & 33.9 & 52.7 & 72.9 &  1.2 &  3.3 &  4.0 &  2.4 &  2.8 \\
Step1X-Edit            & 30.3 & 12.6 & 74.9 &  0.0 &  2.2 &  2.0 &  3.5 &  1.9 \\
OmniGen                & 25.1 & 41.5 & 73.5 &  1.2 &  1.0 &  0.0 &  1.2 &  0.8 \\
EMU2                   & 22.6 & 38.2 & 78.3 &  1.2 &  1.1 &  0.0 &  0.0 &  0.5 \\

\midrule
\rowcolor{gray!20}
BAGEL  & 36.5 & 53.5 & 73.0 & 2.4 & 5.6 & 14.0 & 1.2 & 6.1 \\
\rowcolor{gray!3}
{\scriptsize\hspace{2pt} + Unified Thinker (Qwen2.5-VL-7B)} & 53.3 & 73.6 & 78.1 & 14.1 & 17.7 & 18.0 & 3.5 & 13.6 \\
\rowcolor{gray!5}
{\scriptsize\hspace{2pt} + Unified Thinker (Qwen3-VL-8B)} & 58.7 & 75.7 & 80.9 & 15.2 & 17.7 & 20.0 & 8.2 & 15.5 \\

\rowcolor{gray!20}
Qwen-Image-Edit & 37.2 & 66.4 & 86.9 &  4.7 & 10.0 & 17.0 &  2.4 &  8.9 \\
\rowcolor{gray!3}
{\scriptsize\hspace{2pt} + Unified Thinker (Qwen2.5-VL-7B)}
 & 58.6 & 75.9 & 90.1 & 24.7 & 22.2 & 38.0 &  9.4 & 24.2 \\
\rowcolor{gray!5}
{\scriptsize\hspace{2pt} + Unified Thinker (Qwen3-VL-8B)} &
\textbf{61.9} & \textbf{76.2} & \textbf{90.5} &
\textbf{32.9} & \textbf{30.0} & \textbf{41.0} & \textbf{9.4} & \textbf{28.9} \\

\bottomrule
\end{tabular}
\end{table*}

\begin{table}[th]
\centering
\caption{Results on GEditBench for general instruction-based image editing. We report G\_SC, G\_PQ, and G\_O on the English split.}
\label{tab:gedit}
\setlength{\tabcolsep}{3.5pt} 
\footnotesize 
\renewcommand{\arraystretch}{1.2} 
\begin{tabular}{lccc}
\toprule
\textbf{Model} & \textbf{G\_SC $\uparrow$} & \textbf{G\_PQ $\uparrow$} & \textbf{G\_O $\uparrow$} \\
\midrule
UniWorld-V2                  & 8.29 & 8.02 &7.83 \\
Step1x-edit-v1p2(reflection)& 8.18 & 7.85 & 7.58 \\
Step1x-edit-v1p2(thinking)  & 8.02 & 7.64 & 7.36 \\
Step1X-edit-v1.1             & 7.66 & 7.35 & 6.97 \\
Flux-Kontext-dev             & 7.16 & 7.37 & 6.51 \\
OmniGen2                     &7.16 & 6.77 & 6.41 \\
OmniGen                      & 5.96 & 5.89 & 5.06 \\
AnyEdit                      &3.18 & 5.82 & 3.21 \\

\midrule
\rowcolor{gray!20}
BAGEL & 7.36	& 6.83 & 6.52 \\
\rowcolor{gray!3}
{\scriptsize\hspace{2pt} + Unified Thinker (Qwen2.5-VL-7B)} & 7.29 & 6.88 & 6.53 \\
\rowcolor{gray!3}
{\scriptsize\hspace{2pt} + Unified Thinker (Qwen3-VL-8B)} & 7.38 & 6.75 & 6.60 \\

\rowcolor{gray!20}
Qwen-Image-Edit         & 8.00 & 7.86 & 7.56 \\
\rowcolor{gray!3}
{\scriptsize\hspace{2pt} + Unified Thinker (Qwen2.5-VL-7B)}
 & \textbf{8.17} & 7.94 & 7.67 \\
\rowcolor{gray!3}
{\scriptsize\hspace{2pt} + Unified Thinker (Qwen3-VL-8B)} & 8.15 & \textbf{8.04} & \textbf{7.71} \\

\bottomrule
\end{tabular}
\end{table}

\begin{table}[t]
\centering
\caption{Results on PRISM for general text-to-image generation. We report alignment (Aln), aesthetics (Aes), and average (Avg) using GPT-4.1 as evaluation.}
\label{tab:prism}
\setlength{\tabcolsep}{6pt}
\small
\renewcommand{\arraystretch}{1.2} 
\begin{tabular}{lccc}
\toprule
\textbf{Model} & \textbf{Aln $\uparrow$} & \textbf{Aes $\uparrow$} & \textbf{Avg $\uparrow$} \\
\midrule
Gemini-2.5-Flash-Image & 87.1 & 83.4 & 85.3 \\
Qwen-Image             & 81.1 & 78.6 & 79.9 \\
SEEDream 3.0           & 80.5 & 78.7 & 79.6 \\
HiDream-I1-Full        & 76.1 & 75.6 & 75.9 \\
FLUX.1-Krea-dev        & 74.3 & 75.1 & 74.7 \\
SD3.5-Large            & 73.9 & 73.5 & 73.7 \\
FLUX.1-dev             & 72.4 & 74.9 & 73.7 \\
HiDream-I1-Dev         & 70.3 & 70.0 & 70.2 \\
\midrule
\rowcolor{gray!20}
BAGEL & 66.7 &	63.4 &	65.1 \\
\rowcolor{gray!3}
{\scriptsize\hspace{2pt} + Unified Thinker (Qwen2.5-VL-7B)} & 73.5& 67.7 & 70.6 \\
\rowcolor{gray!3}
{\scriptsize\hspace{2pt} + Unified Thinker (Qwen3-VL-8B)} & 75.1 & 69.2& 72.1 \\
\rowcolor{gray!20}
Qwen-Image-Edit                & 76.9 & 70.7 & 73.8 \\
\rowcolor{gray!3}
{\scriptsize\hspace{2pt} + Unified Thinker (Qwen2.5-VL-7B)}     & 77.3 & \textbf{73.8} & 75.6 \\
\rowcolor{gray!3}
{\scriptsize\hspace{2pt} + Unified Thinker (Qwen3-VL-8B)} & \textbf{83.2} & 73.0 & \textbf{78.1} \\
\bottomrule
\end{tabular}
\end{table}

\begin{table*}[t]
\centering
\caption{Results on WiseBench for reasoning-based text-to-image generation. We report accuracy across six knowledge domains and the overall score.}
\label{tab:wisebench}
\fontsize{10pt}{12pt}\selectfont  
\setlength{\tabcolsep}{7pt}
\begin{tabular}{lccccccl}
\toprule
Model & Cultural & Time & Space & Biology & Physics & Chemistry & Overall \\
\midrule
GPT-4o          & 0.81 & 0.71 & 0.89 & 0.83 & 0.79 & 0.74 & 0.80 \\
Qwen-Image      & 0.62 & 0.63 & 0.77 & 0.57 & 0.75 & 0.40 & 0.62 \\
UniWorld-V2     & 0.60 & 0.61 & 0.70 & 0.53 & 0.64 & 0.32 & 0.58 \\
UniWorld-V1     & 0.53 & 0.55 & 0.73 & 0.45 & 0.59 & 0.41 & 0.55 \\
Manzano-3B      & 0.42 & 0.51 & 0.59 & 0.45 & 0.51 & 0.32 & 0.46 \\
Manzano-30B     & 0.58 & 0.50 & 0.65 & 0.50 & 0.55 & 0.32 & 0.54 \\
OpenUni-B-512   & 0.37 & 0.45 & 0.58 & 0.39 & 0.50 & 0.30 & 0.43 \\
OpenUni-L-512   & 0.51 & 0.49 & 0.64 & 0.48 & 0.63 & 0.35 & 0.52 \\
OpenUni-L-1024  & 0.49 & 0.53 & 0.69 & 0.49 & 0.56 & 0.39 & 0.52 \\
MetaQuery-XL    & 0.56 & 0.55 & 0.62 & 0.49 & 0.63 & 0.41 & 0.55 \\
Liquid          & 0.38 & 0.42 & 0.53 & 0.36 & 0.47 & 0.30 & 0.41 \\
\midrule
\rowcolor{gray!20}
BAGEL  & 0.44 & 0.55 & 0.68 & 0.44 & 0.60 & 0.39 & 0.52 \\
\rowcolor{gray!3}
{\scriptsize\hspace{2pt} + Unified Thinker (Qwen2.5-VL-7B)} & 0.72 & 0.65 & 0.75 & 0.64 & 0.75 & 0.61 & 0.70 \\
\rowcolor{gray!3}
{\scriptsize\hspace{2pt} + Unified Thinker (Qwen3-VL-8B)} & 0.70 & 0.65 & 0.73 & 0.62 & 0.73 & 0.55 & 0.68 \\

\rowcolor{gray!20}
Qwen-Image-Edit                          & 0.62 & 0.63 & 0.77 & 0.57 & 0.75 & 0.40 & 0.62 \\
\rowcolor{gray!3}
{\scriptsize\hspace{2pt} + Unified Thinker (Qwen2.5-VL-7B)}     & 0.75 & 0.66 & 0.78 & 0.75 & 0.79 & 0.61 & 0.73 \\
\rowcolor{gray!3}
{\scriptsize\hspace{2pt} + Unified Thinker (Qwen3-VL-8B)} &
\textbf{0.75} & \textbf{0.70} & \textbf{0.81} & \textbf{0.73} & \textbf{0.81} & \textbf{0.55} & 
\textbf{0.74} \\

\bottomrule
\end{tabular}
\end{table*}

\section{Experiments}

We evaluate Unified Thinker in four settings: general instruction-driven image editing, general text-to-image generation, reasoning-intensive image editing, and reasoning-intensive text-to-image generation. Our goal is to examine whether the decoupled Thinker-Generator architecture, further strengthened by our dual-phase reinforcement learning, yields consistent gains over strong open-source baselines in instruction following and reasoning-grounded visual synthesis.

\subsection{Experimental Setup}

\textbf{Model configuration.} Unless otherwise specified, Unified Thinker uses Qwen2.5-VL-7B, and we additionally report results with Qwen3-VL-8B~\cite{bai2024qwenvl,bai2025qwen3vltechnicalreport}.We use Qwen-Image-Edit~\cite{qwenimage-edit} as the base generator to execute the visual specifications produced by the Thinker. For reinforcement learning and automated evaluation, we adopt Qwen3-VL-30B~\cite{bai2025qwen3vltechnicalreport} as the reward model, which provides feedback on both visual correctness and logical consistency.

\textbf{Training data and setup.} For the supervised cold start, we jointly fine-tune on HieraReason-40K. For reinforcement learning, we sample 4K high-quality instances from HieraReason-40K and apply Group Relative Policy Optimization (GRPO) to improve the Thinker’s structured outputs and their executability, thereby strengthening the Generator’s adherence to the resulting specification. Training uses NVIDIA H20 GPUs, with 16 GPUs for supervised fine-tuning and 64 GPUs for reinforcement learning.

\textbf{Evaluation benchmarks.} We evaluate on WiseBench~\cite{wisebench}, RISEBench~\cite{risebench}, GEditBench~\cite{liu2025step1x-edit}, and PRISMBench~\cite{fang2025fluxreason}. These benchmarks cover diverse knowledge domains and editing operations, requiring models to combine high-level semantic reasoning (e.g., temporal, and logical inference) with low-level visual manipulation (e.g., content preservation).

\subsection{Main Results}

\textbf{Reasoning image editing (RISEBench).} As shown in Table~\ref{tab:risebench}, our method markedly improves reasoning-heavy editing over the base Qwen-Image-Edit and a naive MLLM-thinker baseline. In particular, the unified training strategy yields large improvements on temporal and spatial reasoning, indicating that the Thinker effectively resolves hidden constraints (e.g., temporal shifts or relational edits) and reduces semantic drift during diffusion execution.

\textbf{Reasoning text-to-image (WiseBench).} Table~\ref{tab:wisebench} shows that Unified Thinker achieves the strongest overall performance among open-source models and improves most domain categories, substantially narrowing the gap to closed-source frontier models such as GPT-4o. Gains are especially notable in categories that demand precise entity grounding and knowledge retrieval (e.g., cultural and biology), suggesting that explicit planning helps translate implicit constraints into executable visual specifications.

\textbf{General generation and editing.} Beyond reasoning-centric benchmarks, we further confirm that incorporating the Thinker does not compromise general-purpose generation or editing performance. On PRISM (Table~\ref{tab:prism}), our method achieves a consistent improvement in overall quality, with gains that are mainly reflected in aesthetic preference while preserving prompt-image alignment. On GEditBench (Table~\ref{tab:gedit}), Unified Thinker also delivers modest yet consistent gains across all reported metrics. Together, these results suggest that the planning stage improves instruction decomposition and visual target specification without weakening the Generator’s core rendering ability, and can even provide small benefits under standard, non-reasoning workloads.

\subsection{Ablation Study}

\textbf{Training stage ablation.}
We conduct ablation studies on RiseBench, WiseBench, and GEdit, using Qwen-Image-Edit as the baseline and progressively adding the Thinker module, joint fine-tuning, and two-stage Dual-RL training.

Table~\ref{tab:ablation} shows that introducing the Thinker notably improves performance on reasoning-oriented benchmarks, while slightly hurting low-level editing quality on GEdit, revealing a mild objective trade-off. Joint fine-tuning alleviates this mismatch and stabilizes multi-task behavior, and the proposed two-stage Dual-RL further yields consistent gains across all benchmarks, leading to the best overall results by better aligning reasoning with final visual outcomes.

\textbf{Thinker backbone ablation.} We instantiate Unified Thinker with two backbones(Qwen2.5-VL-7B and Qwen3-VL-8B). Overall, a stronger Thinker backbone tends to yield better reasoning-oriented performance and improves overall editing fidelity on reasoning tasks, whereas the 7B variant can be slightly preferred on PRISM in terms of aesthetics, suggesting a trade-off between logic alignment and visual preference. Moreover, Table~\ref{tab:thinker_ablation_risebench_compact} shows that using an external Thinker (regardless of the specific backbone) consistently outperforms the Qwen-Image-Edit baseline, with most gains coming from improved reasoning and  consistency while visual quality remains comparable.


\begin{table}[t]
\centering
\small
\setlength{\tabcolsep}{6pt}
\caption{Training stage ablation results on RiseBench, WiseBench, and GEdit. The baseline is based on Qwen-Image-Edit. The Thinker is implemented with Qwen2.5-VL-7B and further trained in our framework.}
\fontsize{10pt}{12pt}\selectfont  
\begin{tabular}{lccc}
\toprule
\textbf{Ablation} & \textbf{Rise} $\uparrow$ & \textbf{Wise} $\uparrow$ & \textbf{GEdit} $\uparrow$ \\
\midrule
\rowcolor{gray!15}
baseline & 8.9  & 0.62 & 7.56 \\
{\scriptsize\hspace{2pt} + Thinker }& 16.4 & 0.66 & 7.49 \\
{\scriptsize\hspace{2pt} + Joint fine-tune }& 20.2 & 0.68 & 7.52 \\
{\scriptsize\hspace{2pt} +Dual-RL stage 1 }& 21.9 & 0.72 & 7.61 \\
{\scriptsize\hspace{2pt} + Dual-RL stage 2} & 24.2 & 0.73 & 7.67 \\
\bottomrule
\end{tabular}

\label{tab:ablation}
\end{table}


\begin{table}[t]
\centering
\caption{Ablation of the Thinker design on RiseBench. We report Reason., Consist., Visual., and \textit{Overall}, where \textit{Overall} is the average accuracy over Temporal, Causal, Spatial, and Logical. The baseline is Qwen-Image-Edit.}
\label{tab:thinker_ablation_risebench_compact}
\setlength{\tabcolsep}{2pt}
\fontsize{10pt}{12pt}\selectfont  
\begin{tabular}{lcccc}
\toprule
\textbf{Model} & \textbf{Reason.} & \textbf{Consist.} & \textbf{Visual.} & \textbf{Overall} \\
\midrule
\rowcolor{gray!15}
baseline & 37.2 & 66.4 & 86.9 & 8.9 \\
{\scriptsize\hspace{2pt} + Gemini-2.5-Pro} & 64.3 & 71.9 & 88.3 & 25.2 \\
{\scriptsize\hspace{2pt} + GPT-5} & 67.4 & 76.6 & 86.3 & 26.9 \\
{\scriptsize\hspace{2pt} + Qwen3-VL-30B} & 57.6 & 75.9 & 86.6 & 23.1 \\
{\scriptsize\hspace{2pt} + Unified Thinker (7B) } & 58.6 & 75.9 & 90.1 & 24.2 \\
\bottomrule
\end{tabular}
\end{table}

\textbf{Cross-generator ablation.}
\label{sec:thinker_transfer}
We evaluate the transferability of the Thinker module by applying the Unified Thinker trained with the Qwen-Image-Edit pipeline to a different generator (BAGEL). As shown in Tables~\ref{tab:risebench}, \ref{tab:wisebench}, and~\ref{tab:gedit}, adding Thinker consistently improves BAGEL on both RiseBench and GEditBench, demonstrating that the module generalizes beyond the training generator and can be integrated into other generation models with stable gains.

\section{Conclusion}


We propose \textsc{Unified Thinker}, a decoupled Thinker–Generator framework that equips diffusion models with transferable reasoning and planning. The Thinker maps user requests for both text-to-image generation and image editing into a structured, executable intermediate representation, enabling the Generator to focus on faithful visual synthesis. We build a $\sim$40K cold-start training corpus with strict formatting and further enhance both planning and execution with a two-stage RL pipeline. Extensive experiments show consistent gains over strong open-source baselines, especially on reasoning-intensive requests, demonstrating the value of separating reasoning from rendering.

\section{Limitations}

Our approach still depends on the quality and coverage of the intermediate representation, training data, and automatic rewards used during RL, which can introduce bias and limit generalization beyond the evaluated benchmarks. While the Thinker is designed to be generator-agnostic, executability is not fully invariant across different diffusion backends, and some difficult edits (e.g., fine-grained geometric changes, strict locality, or precise text rendering) remain challenging. Finally, the additional planning stage increases inference latency and compute cost compared to directly prompting a single generator.

\bibliography{custom}

@inproceedings{fu2024guiding,
  title={Guiding Instruction-based Image Editing via Multimodal Large Language Models},
  author={Fu, Tsu-Jui and Hsu, Wenze and Wang, William Yang and Li, Shang-Hua and Cohen, Scott and Wang, Yang},
  booktitle={International Conference on Learning Representations (ICLR)},
  year={2024}
}

@misc{bai2024qwenvl,
      title={Qwen2.5-VL Technical Report}, 
      author={Shuai Bai and Keqin Chen and Xuejing Liu and Jialin Wang and Wenbin Ge and Sibo Song and Kai Dang and Peng Wang and Shijie Wang and Jun Tang and Humen Zhong and Yuanzhi Zhu and Mingkun Yang and Zhaohai Li and Jianqiang Wan and Pengfei Wang and Wei Ding and Zheren Fu and Yiheng Xu and Jiabo Ye and Xi Zhang and Tianbao Xie and Zesen Cheng and Hang Zhang and Zhibo Yang and Haiyang Xu and Junyang Lin},
      year={2025},
      eprint={2502.13923},
      archivePrefix={arXiv},
      primaryClass={cs.CV},
}

@inproceedings{xu2024diffusion_dpo,
  author       = {Rafael Rafailov and
                  Archit Sharma and
                  Eric Mitchell and
                  Christopher D. Manning and
                  Stefano Ermon and
                  Chelsea Finn},
  editor       = {Alice Oh and
                  Tristan Naumann and
                  Amir Globerson and
                  Kate Saenko and
                  Moritz Hardt and
                  Sergey Levine},
  title        = {Direct Preference Optimization: Your Language Model is Secretly a
                  Reward Model},
  booktitle    = {Advances in Neural Information Processing Systems 36: Annual Conference
                  on Neural Information Processing Systems 2023, NeurIPS 2023, New Orleans,
                  LA, USA, December 10 - 16, 2023},
  year         = {2023},
  bibsource    = {dblp computer science bibliography, https://dblp.org}
}

@article{dhariwal2021diffusion,
  title={Diffusion Models Beat GANs on Image Synthesis},
  author={Dhariwal, Prafulla and Nichol, Alex},
  journal={arXiv preprint arXiv:2105.05233},
  year={2021}
}

@inproceedings{li2024smartedit,
  author       = {Yuzhou Huang and
                  Liangbin Xie and
                  Xintao Wang and
                  Ziyang Yuan and
                  Xiaodong Cun and
                  Yixiao Ge and
                  Jiantao Zhou and
                  Chao Dong and
                  Rui Huang and
                  Ruimao Zhang and
                  Ying Shan},
  title        = {SmartEdit: Exploring Complex Instruction-Based Image Editing with
                  Multimodal Large Language Models},
  booktitle    = {{IEEE/CVF} Conference on Computer Vision and Pattern Recognition,
                  {CVPR} 2024, Seattle, WA, USA, June 16-22, 2024},
  pages        = {8362--8371},
  publisher    = {{IEEE}},
  year         = {2024},
  doi          = {10.1109/CVPR52733.2024.00799},
  bibsource    = {dblp computer science bibliography, https://dblp.org}
}

@article{ho2020denoising,

  title={Denoising diffusion probabilistic models},

  author={Ho, Jonathan and Jain, Ajay and Abbeel, Pieter},

  journal={Advances in neural information processing systems},

  year={2020}

}

@inproceedings{peebles2023scalable,
  title={Scalable diffusion models with transformers},
  author={Peebles, William and Xie, Saining},
  booktitle={Proceedings of the IEEE/CVF international conference on computer vision},
  pages={4195--4205},
  year={2023}
}

@article{flowmatch,
  title={Flow matching for generative modeling},
  author={Lipman, Yaron and Chen, Ricky TQ and Ben-Hamu, Heli and Nickel, Maximilian and Le, Matt},
  journal={arXiv preprint arXiv:2210.02747},
  year={2022}
}

@inproceedings{sd3.5,
  author= {Patrick Esser and
                  Sumith Kulal and
                  Andreas Blattmann and
                  Rahim Entezari and
                  Jonas M{\"{u}}ller and
                  Harry Saini and
                  Yam Levi and
                  Dominik Lorenz and
                  Axel Sauer and
                  Frederic Boesel and
                  Dustin Podell and
                  Tim Dockhorn and
                  Zion English and
                  Robin Rombach},
  title        = {Scaling Rectified Flow Transformers for High-Resolution Image Synthesis},
  booktitle    = {Forty-first International Conference on Machine Learning, {ICML} 2024,
                  Vienna, Austria, July 21-27, 2024},
  publisher    = {OpenReview.net},
  year         = {2024},
  bibsource    = {dblp computer science bibliography, https://dblp.org}
}

@article{bagel2025,
  title   = {Emerging Properties in Unified Multimodal Pretraining},
  author  = {Deng, Chaorui and Zhu, Deyao and Li, Kunchang and Gou, Chenhui and Li, Feng and Wang, Zeyu and Zhong, Shu and Yu, Weihao and Nie, Xiaonan and Song, Ziang and Shi, Guang and Fan, Haoqi},
  journal = {arXiv preprint arXiv:2505.14683},
  year    = {2025}
}

@article{show-o2,
  title={Show-o2: Improved Native Unified Multimodal Models},
  author={Xie, Jinheng and Yang, Zhenheng and Shou, Mike Zheng},
  journal={arXiv preprint arXiv:2506.15564},
  year={2025}
}

@inproceedings{janus,
  author       = {Chengyue Wu and
                  Xiaokang Chen and
                  Zhiyu Wu and
                  Yiyang Ma and
                  Xingchao Liu and
                  Zizheng Pan and
                  Wen Liu and
                  Zhenda Xie and
                  Xingkai Yu and
                  Chong Ruan and
                  Ping Luo},
  title        = {Janus: Decoupling Visual Encoding for Unified Multimodal Understanding
                  and Generation},
  booktitle    = {{IEEE/CVF} Conference on Computer Vision and Pattern Recognition,
                  {CVPR} 2025, Nashville, TN, USA, June 11-15, 2025},
  pages        = {12966--12977},
  publisher    = {Computer Vision Foundation / {IEEE}},
  year         = {2025},
  doi          = {10.1109/CVPR52734.2025.01210},
  bibsource    = {dblp computer science bibliography, https://dblp.org}
}

@article{qwenimage-edit,
      title={Qwen-Image Technical Report}, 
      author={Chenfei Wu and Jiahao Li and Jingren Zhou and Junyang Lin and Kaiyuan Gao and Kun Yan and Sheng-ming Yin and Shuai Bai and Xiao Xu and Yilei Chen and Yuxiang Chen and Zecheng Tang and Zekai Zhang and Zhengyi Wang and An Yang and Bowen Yu and Chen Cheng and Dayiheng Liu and Deqing Li and Hang Zhang and Hao Meng and Hu Wei and Jingyuan Ni and Kai Chen and Kuan Cao and Liang Peng and Lin Qu and Minggang Wu and Peng Wang and Shuting Yu and Tingkun Wen and Wensen Feng and Xiaoxiao Xu and Yi Wang and Yichang Zhang and Yongqiang Zhu and Yujia Wu and Yuxuan Cai and Zenan Liu},
      year={2025},
      eprint={2508.02324},
      archivePrefix={arXiv},
      primaryClass={cs.CV},
}

@inproceedings{brooks2023instructpix2pix,
  title={Instructpix2pix: Learning to follow image editing instructions},
  author={Brooks, Tim and Holynski, Aleksander and Efros, Alexei A},
  booktitle={Proceedings of the IEEE/CVF conference on computer vision and pattern recognition},
  pages={18392--18402},
  year={2023}
}

@inproceedings{ju2024brushnet,
  title={Brushnet: A plug-and-play image inpainting model with decomposed dual-branch diffusion},
  author={Ju, Xuan and Liu, Xian and Wang, Xintao and Bian, Yuxuan and Shan, Ying and Xu, Qiang},
  booktitle={European Conference on Computer Vision},
  pages={150--168},
  year={2024},
  organization={Springer}
}

@inproceedings{zhuang2024task,
  title={A task is worth one word: Learning with task prompts for high-quality versatile image inpainting},
  author={Zhuang, Junhao and Zeng, Yanhong and Liu, Wenran and Yuan, Chun and Chen, Kai},
  booktitle={European Conference on Computer Vision},
  pages={195--211},
  year={2024},
  organization={Springer}
}

@inproceedings{yu2025anyedit,
  title={Anyedit: Mastering unified high-quality image editing for any idea},
  author={Yu, Qifan and Chow, Wei and Yue, Zhongqi and Pan, Kaihang and Wu, Yang and Wan, Xiaoyang and Li, Juncheng and Tang, Siliang and Zhang, Hanwang and Zhuang, Yueting},
  booktitle={Proceedings of the Computer Vision and Pattern Recognition Conference},
  pages={26125--26135},
  year={2025}
}

@inproceedings{rombach2022high,
  title={High-resolution image synthesis with latent diffusion models},
  author={Rombach, Robin and Blattmann, Andreas and Lorenz, Dominik and Esser, Patrick and Ommer, Bj{\"o}rn},
  booktitle={Proceedings of the IEEE/CVF conference on computer vision and pattern recognition},
  pages={10684--10695},
  year={2022}
}

@article{lin2025uniworld,
  author       = {Bin Lin and
                  Zongjian Li and
                  Xinhua Cheng and
                  Yuwei Niu and
                  Yang Ye and
                  Xianyi He and
                  Shenghai Yuan and
                  Wangbo Yu and
                  Shaodong Wang and
                  Yunyang Ge and
                  Yatian Pang and
                  Li Yuan},
  title        = {UniWorld-V1: High-Resolution Semantic Encoders for Unified Visual
                  Understanding and Generation},
  journal      = {CoRR},
  volume       = {abs/2506.03147},
  year         = {2025},
  doi          = {10.48550/ARXIV.2506.03147},
  eprinttype    = {arXiv},
  eprint       = {2506.03147},
  bibsource    = {dblp computer science bibliography, https://dblp.org}
}

@article{t2ir1,
  title={T2i-r1: Reinforcing image generation with collaborative semantic-level and token-level cot},
  author={Jiang, Dongzhi and Guo, Ziyu and Zhang, Renrui and Zong, Zhuofan and Li, Hao and Zhuo, Le and Yan, Shilin and Heng, Pheng-Ann and Li, Hongsheng},
  journal={arXiv preprint arXiv:2505.00703},
  year={2025}
}

@article{liao2025imagegen-cot,
  title={Imagegen-cot: Enhancing text-to-image in-context learning with chain-of-thought reasoning},
  author={Liao, Jiaqi and Yang, Zhengyuan and Li, Linjie and Li, Dianqi and Lin, Kevin and Cheng, Yu and Wang, Lijuan},
  journal={arXiv preprint arXiv:2503.19312},
  year={2025}
}

@article{mi2025thinkdiff,
  title={I Think, Therefore I Diffuse: Enabling Multimodal In-Context Reasoning in Diffusion Models},
  author={Mi, Zhenxing and Wang, Kuan-Chieh and Qian, Guocheng and Ye, Hanrong and Liu, Runtao and Tulyakov, Sergey and Aberman, Kfir and Xu, Dan},
  journal={arXiv preprint arXiv:2502.10458},
  year={2025}
}

@article{rgenie,
  title={R-Genie: Reasoning-Guided Generative Image Editing},
  author={Zhang, Dong and He, Lingfeng and Yan, Rui and Shen, Fei and Tang, Jinhui},
  journal={arXiv preprint arXiv:2505.17768},
  year={2025}
}

@article{wu2025omnigen2,
  author       = {Chenyuan Wu and
                  Pengfei Zheng and
                  Ruiran Yan and
                  Shitao Xiao and
                  Xin Luo and
                  Yueze Wang and
                  Wanli Li and
                  Xiyan Jiang and
                  Yexin Liu and
                  Junjie Zhou and
                  Ze Liu and
                  Ziyi Xia and
                  Chaofan Li and
                  Haoge Deng and
                  Jiahao Wang and
                  Kun Luo and
                  Bo Zhang and
                  Defu Lian and
                  Xinlong Wang and
                  Zhongyuan Wang and
                  Tiejun Huang and
                  Zheng Liu},
  title        = {OmniGen2: Exploration to Advanced Multimodal Generation},
  journal      = {CoRR},
  volume       = {abs/2506.18871},
  year         = {2025},
  doi          = {10.48550/ARXIV.2506.18871},
  eprinttype    = {arXiv},
  eprint       = {2506.18871},
  bibsource    = {dblp computer science bibliography, https://dblp.org}
}

@article{li2025reflectdit,
  title={Reflect-DiT: Inference-Time Scaling for Text-to-Image Diffusion Transformers via In-Context Reflection},
  author={Li, Shufan and Kallidromitis, Konstantinos and Gokul, Akash and Koneru, Arsh and Kato, Yusuke and Kozuka, Kazuki and Grover, Aditya},
  journal={arXiv preprint arXiv:2503.12271},
  year={2025}
}

@article{li2025editthinker,
      title={EditThinker: Unlocking Iterative Reasoning for Any Image Editor}, 
      author={Hongyu Li and Manyuan Zhang and Dian Zheng and Ziyu Guo and Yimeng Jia and Kaituo Feng and Hao Yu and Yexin Liu and Yan Feng and Peng Pei and Xunliang Cai and Linjiang Huang and Hongsheng Li and Si Liu},
      year={2025},
      eprint={2512.05965},
      archivePrefix={arXiv},
      primaryClass={cs.CV},
}

@article{unicot,
  title={Uni-cot: Towards unified chain-of-thought reasoning across text and vision},
  author={Qin, Luozheng and Gong, Jia and Sun, Yuqing and Li, Tianjiao and Yang, Mengping and Yang, Xiaomeng and Qu, Chao and Tan, Zhiyu and Li, Hao},
  journal={arXiv preprint arXiv:2508.05606},
  year={2025}
}

@article{wang2025mint,
  author       = {Yi Wang and
                  Mushui Liu and
                  Wanggui He and
                  Longxiang Zhang and
                  Ziwei Huang and
                  Guanghao Zhang and
                  Fangxun Shu and
                  Tao Zhong and
                  Dong She and
                  Zhelun Yu and
                  Haoyuan Li and
                  Weilong Dai and
                  Mingli Song and
                  Jie Song and
                  Hao Jiang},
  title        = {{MINT:} Multi-modal Chain of Thought in Unified Generative Models
                  for Enhanced Image Generation},
  journal      = {CoRR},
  volume       = {abs/2503.01298},
  year         = {2025},
  doi          = {10.48550/ARXIV.2503.01298},
  eprinttype    = {arXiv},
  eprint       = {2503.01298},
  bibsource    = {dblp computer science bibliography, https://dblp.org}
}

@article{huang2025interleaving,
        author = {Wenxuan Huang and
                  Shuang Chen and
                  Zheyong Xie and
                  Shaosheng Cao and
                  Shixiang Tang and
                  Yufan Shen and
                  Qingyu Yin and
                  Wenbo Hu and
                  Xiaoman Wang and
                  Yuntian Tang and
                  Junbo Qiao and
                  Yue Guo and
                  Yao Hu and
                  Zhenfei Yin and
                  Philip Torr and
                  Yu Cheng and
                  Wanli Ouyang and
                  Shaohui Lin},
  title        = {Interleaving Reasoning for Better Text-to-Image Generation},
  journal      = {CoRR},
  volume       = {abs/2509.06945},
  year         = {2025},
  doi          = {10.48550/ARXIV.2509.06945},
  eprinttype    = {arXiv},
  eprint       = {2509.06945},
  bibsource    = {dblp computer science bibliography, https://dblp.org}
}

@article{reason2edit,
  title={Reasoning to Edit: Hypothetical Instruction-Based Image Editing with Visual Reasoning},
  author={He, Qingdong and Chen, Xueqin and Wang, Chaoyi and Pan, Yanjie and Hu, Xiaobin and Gan, Zhenye and Wang, Yabiao and Wang, Chengjie and Li, Xiangtai and Zhang, Jiangning},
  journal={arXiv preprint arXiv:2507.01908},
  year={2025}
}

@misc{yin2025reasonedit,
      title={ReasonEdit: Towards Reasoning-Enhanced Image Editing Models}, 
      author={Fukun Yin and Shiyu Liu and Yucheng Han and Zhibo Wang and Peng Xing and Rui Wang and Wei Cheng and Yingming Wang and Aojie Li and Zixin Yin and Pengtao Chen and Xiangyu Zhang and Daxin Jiang and Xianfang Zeng and Gang Yu},
      year={2025},
      eprint={2511.22625},
      archivePrefix={arXiv},
      primaryClass={cs.CV},
}

@article{guo2025canwegen,
  author       = {Ziyu Guo and
                  Renrui Zhang and
                  Chengzhuo Tong and
                  Zhizheng Zhao and
                  Peng Gao and
                  Hongsheng Li and
                  Pheng{-}Ann Heng},
  title        = {Can We Generate Images with CoT? Let's Verify and Reinforce Image
                  Generation Step by Step},
  journal      = {CoRR},
  volume       = {abs/2501.13926},
  year         = {2025},
  doi          = {10.48550/ARXIV.2501.13926},
  eprinttype    = {arXiv},
  eprint       = {2501.13926},
  bibsource    = {dblp computer science bibliography, https://dblp.org}
}

@misc{nanobanana,
      title={Gemini 2.5: Pushing the Frontier with Advanced Reasoning, Multimodality, Long Context, and Next Generation Agentic Capabilities}, 
      author={Gheorghe Comanici and Eric Bieber and Mike Schaekermann and Ice Pasupat and Noveen Sachdeva and Inderjit Dhillon and Marcel Blistein and Ori Ram and Dan Zhang and Evan Rosen and Luke Marris and Sam Petulla and Colin Gaffney and Asaf Aharoni and Nathan Lintz and Tiago Cardal Pais and Henrik Jacobsson and Idan Szpektor and Nan-Jiang Jiang and 3416 others.},
      year={2025},
      eprint={2507.06261},
      archivePrefix={arXiv},
      primaryClass={cs.CL},
}

@article{hurst2024gpt4o,
  author       = {Aaron Hurst and
                  Adam Lerer and
                  Adam P. Goucher and
                  Adam Perelman and
                  Aditya Ramesh and
                  Aidan Clark and
                  AJ Ostrow and
                  Akila Welihinda and
                  Alan Hayes and
                  Alec Radford and
                  Aleksander Madry and
                  Alex Baker{-}Whitcomb and
                  Alex Beutel and
                  Alex Borzunov and
                  Alex Carney and
                  Alex Chow and
                  Alex Kirillov and
                  Alex Nichol and
                  Alex Paino and
                  Alex Renzin and
                  Alex Tachard Passos and
                  Alexander Kirillov and
                  Alexi Christakis and
                  Alexis Conneau and
                  Ali Kamali and
                  Allan Jabri and
                  Allison Moyer and
                  Allison Tam and
                  Amadou Crookes and
                  Amin Tootoonchian and
                  Ananya Kumar and
                  Andrea Vallone and
                  Andrej Karpathy and
                  Andrew Braunstein and
                  Andrew Cann and
                  Andrew Codispoti and
                  Andrew Galu and
                  Andrew Kondrich and
                  Andrew Tulloch and
                  Andrey Mishchenko and
                  Angela Baek and
                  Angela Jiang and
                  Antoine Pelisse and
                  Antonia Woodford and
                  Anuj Gosalia and
                  Arka Dhar and
                  Ashley Pantuliano and
                  Avi Nayak and
                  Avital Oliver and
                  Barret Zoph and
                  Behrooz Ghorbani and
                  Ben Leimberger and
                  Ben Rossen and
                  Ben Sokolowsky and
                  Ben Wang and
                  Benjamin Zweig and
                  Beth Hoover and
                  Blake Samic and
                  Bob McGrew and
                  Bobby Spero and
                  Bogo Giertler and
                  Bowen Cheng and
                  Brad Lightcap and
                  Brandon Walkin and
                  Brendan Quinn and
                  Brian Guarraci and
                  Brian Hsu and
                  Bright Kellogg and
                  Brydon Eastman and
                  Camillo Lugaresi and
                  Carroll L. Wainwright and
                  Cary Bassin and
                  Cary Hudson and
                  Casey Chu and
                  Chad Nelson and
                  Chak Li and
                  Chan Jun Shern and
                  Channing Conger and
                  Charlotte Barette and
                  Chelsea Voss and
                  Chen Ding and
                  Cheng Lu and
                  Chong Zhang and
                  Chris Beaumont and
                  Chris Hallacy and
                  Chris Koch and
                  Christian Gibson and
                  Christina Kim and
                  Christine Choi and
                  Christine McLeavey and
                  Christopher Hesse and
                  Claudia Fischer and
                  Clemens Winter and
                  Coley Czarnecki and
                  Colin Jarvis and
                  Colin Wei and
                  Constantin Koumouzelis and
                  Dane Sherburn},
  title        = {GPT-4o System Card},
  journal      = {CoRR},
  volume       = {abs/2410.21276},
  year         = {2024},
  doi          = {10.48550/ARXIV.2410.21276},
  eprinttype    = {arXiv},
  eprint       = {2410.21276},
  bibsource    = {dblp computer science bibliography, https://dblp.org}
}

@article{risebench,
  title={Envisioning Beyond the Pixels: Benchmarking Reasoning-Informed Visual Editing},
  author={Zhao, Xiangyu and Zhang, Peiyuan and Tang, Kexian and Li, Hao and Zhang, Zicheng and Zhai, Guangtao and Yan, Junchi and Yang, Hua and Yang, Xue and Duan, Haodong},
  journal={arXiv preprint arXiv:2504.02826},
  year={2025}
}

@article{liu2025step1x-edit,
  title={Step1X-Edit: A Practical Framework for General Image Editing}, 
  author={Shiyu Liu and Yucheng Han and Peng Xing and Fukun Yin and Rui Wang and Wei Cheng and Jiaqi Liao and Yingming Wang and Honghao Fu and Chunrui Han and Guopeng Li and Yuang Peng and Quan Sun and Jingwei Wu and Yan Cai and Zheng Ge and Ranchen Ming and Lei Xia and Xianfang Zeng and Yibo Zhu and Binxing Jiao and Xiangyu Zhang and Gang Yu and Daxin Jiang},
  journal={arXiv preprint arXiv:2504.17761},
  year={2025}
}

@article{wisebench,
  title={WISE: A World Knowledge-Informed Semantic Evaluation for Text-to-Image Generation},
  author={Niu, Yuwei and Ning, Munan and Zheng, Mengren and Jin, Weiyang and Lin, Bin and Jin, Peng and Liao, Jiaqi and Ning, Kunpeng and Feng, Chaoran and Zhu, Bin and Yuan, Li},
  journal={arXiv preprint arXiv:2503.07265},
  year={2025}
}

@article{qian2025pico,
  title={Pico-Banana-400K: A Large-Scale Dataset for Text-Guided Image Editing},
  author={Qian, Yusu and Bocek-Rivele, Eli and Song, Liangchen and Tong, Jialing and Yang, Yinfei and Lu, Jiasen and Hu, Wenze and Gan, Zhe},
  journal={arXiv preprint arXiv:2510.19808},
  year={2025}
}

@article{fang2025fluxreason,
      title={FLUX-Reason-6M \& PRISM-Bench: A Million-Scale Text-to-Image Reasoning Dataset and Comprehensive Benchmark}, 
      author={Fang, Rongyao and Yu, Aldrich and Duan, Chengqi and Huang, Linjiang and Bai, Shuai and Cai, Yuxuan and Wang, Kun and Liu, Si and Liu, Xihui and Li, Hongsheng},
      journal={arXiv preprint arXiv:2509.09680},
      year={2025}
}

@article{han2025unireditbench,
  author       = {Feng Han and
                  Yibin Wang and
                  Chenglin Li and
                  Zheming Liang and
                  Dianyi Wang and
                  Yang Jiao and
                  Zhipeng Wei and
                  Chao Gong and
                  Cheng Jin and
                  Jingjing Chen and
                  Jiaqi Wang},
  title        = {UniREditBench: {A} Unified Reasoning-based Image Editing Benchmark},
  journal      = {CoRR},
  volume       = {abs/2511.01295},
  year         = {2025},
  doi          = {10.48550/ARXIV.2511.01295},
  eprinttype    = {arXiv},
  eprint       = {2511.01295},
  bibsource    = {dblp computer science bibliography, https://dblp.org}
}

@misc{bai2025qwen3vltechnicalreport,
      title={Qwen3-VL Technical Report}, 
      author={Shuai Bai and Yuxuan Cai and Ruizhe Chen and Keqin Chen and Xionghui Chen and Zesen Cheng and Lianghao Deng and Wei Ding and Chang Gao and Chunjiang Ge and Wenbin Ge and Zhifang Guo and Qidong Huang and Jie Huang and Fei Huang and Binyuan Hui and Shutong Jiang and Zhaohai Li and Mingsheng Li and Mei Li and Kaixin Li and Zicheng Lin and Junyang Lin and Xuejing Liu and Jiawei Liu and Chenglong Liu and Yang Liu and Dayiheng Liu and Shixuan Liu and Dunjie Lu and Ruilin Luo and Chenxu Lv and Rui Men and Lingchen Meng and Xuancheng Ren and Xingzhang Ren and Sibo Song and Yuchong Sun and Jun Tang and Jianhong Tu and Jianqiang Wan and Peng Wang and Pengfei Wang and Qiuyue Wang and Yuxuan Wang and Tianbao Xie and Yiheng Xu and Haiyang Xu and Jin Xu and Zhibo Yang and Mingkun Yang and Jianxin Yang and An Yang and Bowen Yu and Fei Zhang and Hang Zhang and Xi Zhang and Bo Zheng and Humen Zhong and Jingren Zhou and Fan Zhou and Jing Zhou and Yuanzhi Zhu and Ke Zhu},
      year={2025},
      eprint={2511.21631},
      archivePrefix={arXiv},
      primaryClass={cs.CV},
}

@article{flowgrpo,
  title={Flow-grpo: Training flow matching models via online rl},
  author={Liu, Jie and Liu, Gongye and Liang, Jiajun and Li, Yangguang and Liu, Jiaheng and Wang, Xintao and Wan, Pengfei and Zhang, Di and Ouyang, Wanli},
  journal={arXiv preprint arXiv:2505.05470},
  year={2025}
}

@article{guo2025deepseekr1,
  author= {Daya Guo and
                  Dejian Yang and
                  Haowei Zhang and
                  Junxiao Song and
                  Peiyi Wang and
                  Qihao Zhu and
                  Runxin Xu and
                  Ruoyu Zhang and
                  Shirong Ma and
                  Xiao Bi and
                  Xiaokang Zhang and
                  Xingkai Yu and
                  Yu Wu and
                  Z. F. Wu and
                  Zhibin Gou and
                  Zhihong Shao and
                  Zhuoshu Li and
                  Ziyi Gao and
                  Aixin Liu and
                  Bing Xue and
                  Bingxuan Wang and
                  Bochao Wu and
                  Bei Feng and
                  Chengda Lu and
                  Chenggang Zhao and
                  Chengqi Deng and
                  Chong Ruan and
                  Damai Dai and
                  Deli Chen and
                  Dongjie Ji and
                  Erhang Li and
                  Fangyun Lin and
                  Fucong Dai and
                  Fuli Luo and
                  Guangbo Hao and
                  Guanting Chen and
                  Guowei Li and
                  Hao Zhang and
                  Hanwei Xu and
                  Honghui Ding and
                  Huazuo Gao and
                  Hui Qu and
                  Hui Li and
                  Jianzhong Guo and
                  Jiashi Li and
                  Jingchang Chen and
                  Jingyang Yuan and
                  Jinhao Tu and
                  Junjie Qiu and
                  Junlong Li and
                  J. L. Cai and
                  Jiaqi Ni and
                  Jian Liang and
                  Jin Chen and
                  Kai Dong and
                  Kai Hu and
                  Kaichao You and
                  Kaige Gao and
                  Kang Guan and
                  Kexin Huang and
                  Kuai Yu and
                  Lean Wang and
                  Lecong Zhang and
                  Liang Zhao and
                  Litong Wang and
                  Liyue Zhang and
                  Lei Xu and
                  Leyi Xia and
                  Mingchuan Zhang and
                  Minghua Zhang and
                  Minghui Tang and
                  Mingxu Zhou and
                  Meng Li and
                  Miaojun Wang and
                  Mingming Li and
                  Ning Tian and
                  Panpan Huang and
                  Peng Zhang and
                  Qiancheng Wang and
                  Qinyu Chen and
                  Qiushi Du and
                  Ruiqi Ge and
                  Ruisong Zhang and
                  Ruizhe Pan and
                  Runji Wang and
                  R. J. Chen and
                  R. L. Jin and
                  Ruyi Chen and
                  Shanghao Lu and
                  Shangyan Zhou and
                  Shanhuang Chen and
                  Shengfeng Ye and
                  Shiyu Wang and
                  Shuiping Yu and
                  Shunfeng Zhou and
                  Shuting Pan and
                  S. S. Li and
                  Shuang Zhou and
                  Shaoqing Wu and
                  Tao Yun and
                  Tian Pei and
                  Tianyu Sun and
                  Tao Wang and
                  Wangding Zeng and
                  Wen Liu and
                  Wenfeng Liang and
                  Wenjun Gao and
                  Wenqin Yu and
                  Wentao Zhang and
                  W. L. Xiao and
                  Wei An and
                  Xiaodong Liu and
                  Xiaohan Wang and
                  Xiaokang Chen and
                  Xiaotao Nie and
                  Xin Cheng and
                  Xin Liu and
                  Xin Xie and
                  Xingchao Liu and
                  Xinyu Yang and
                  Xinyuan Li and
                  Xuecheng Su and
                  Xuheng Lin and
                  X. Q. Li and
                  Xiangyue Jin and
                  Xiaojin Shen and
                  Xiaosha Chen and
                  Xiaowen Sun and
                  Xiaoxiang Wang and
                  Xinnan Song and
                  Xinyi Zhou and
                  Xianzu Wang and
                  Xinxia Shan and
                  Y. K. Li and
                  Y. Q. Wang and
                  Y. X. Wei and
                  Yang Zhang and
                  Yanhong Xu and
                  Yao Li and
                  Yao Zhao and
                  Yaofeng Sun and
                  Yaohui Wang and
                  Yi Yu and
                  Yichao Zhang and
                  Yifan Shi and
                  Yiliang Xiong and
                  Ying He and
                  Yishi Piao and
                  Yisong Wang and
                  Yixuan Tan and
                  Yiyang Ma and
                  Yiyuan Liu and
                  Yongqiang Guo and
                  Yuan Ou and
                  Yuduan Wang and
                  Yue Gong and
                  Yuheng Zou and
                  Yujia He and
                  Yunfan Xiong and
                  Yuxiang Luo and
                  Yuxiang You and
                  Yuxuan Liu and
                  Yuyang Zhou and
                  Y. X. Zhu and
                  Yanping Huang and
                  Yaohui Li and
                  Yi Zheng and
                  Yuchen Zhu and
                  Yunxian Ma and
                  Ying Tang and
                  Yukun Zha and
                  Yuting Yan and
                  Z. Z. Ren and
                  Zehui Ren and
                  Zhangli Sha and
                  Zhe Fu and
                  Zhean Xu and
                  Zhenda Xie and
                  Zhengyan Zhang and
                  Zhewen Hao and
                  Zhicheng Ma and
                  Zhigang Yan and
                  Zhiyu Wu and
                  Zihui Gu and
                  Zijia Zhu and
                  Zijun Liu and
                  Zilin Li and
                  Ziwei Xie and
                  Ziyang Song and
                  Zizheng Pan and
                  Zhen Huang and
                  Zhipeng Xu and
                  Zhongyu Zhang and
                  Zhen Zhang},
  title        = {DeepSeek-R1 incentivizes reasoning in LLMs through reinforcement learning},
  journal      = {Nat.},
  volume       = {645},
  number       = {8081},
  pages        = {633--638},
  year         = {2025},
  doi          = {10.1038/S41586-025-09422-Z},
  bibsource    = {dblp computer science bibliography, https://dblp.org}
}

@article{blackforest2024flux,
  title={FLUX.1: Text-to-Image Synthesis via Flow Matching},
  author={Black Forest Labs},
  journal={Technical Announcement},
  year={2024},
}

@article{xiao2024omnigen,
  author       = {Shitao Xiao and
                  Yueze Wang and
                  Junjie Zhou and
                  Huaying Yuan and
                  Xingrun Xing and
                  Ruiran Yan and
                  Chaofan Li and
                  Shuting Wang and
                  Tiejun Huang and
                  Zheng Liu},
  title        = {OmniGen: Unified Image Generation},
  booktitle    = {{IEEE/CVF} Conference on Computer Vision and Pattern Recognition,
                  {CVPR} 2025, Nashville, TN, USA, June 11-15, 2025},
  pages        = {13294--13304},
  publisher    = {Computer Vision Foundation / {IEEE}},
  year         = {2025},
  doi          = {10.1109/CVPR52734.2025.01241},
  bibsource    = {dblp computer science bibliography, https://dblp.org}
}

@misc{emu2,
      title={Generative Multimodal Models are In-Context Learners}, 
      author={Quan Sun and Yufeng Cui and Xiaosong Zhang and Fan Zhang and Qiying Yu and Zhengxiong Luo and Yueze Wang and Yongming Rao and Jingjing Liu and Tiejun Huang and Xinlong Wang},
      year={2024},
      eprint={2312.13286},
      archivePrefix={arXiv},
      primaryClass={cs.CV},
}

@article{ovis,
  title={Ovis: Structural Embedding Alignment for Multimodal Large Language Model}, 
  author={Shiyin Lu and Yang Li and Qing-Guo Chen and Zhao Xu and Weihua Luo and Kaifu Zhang and Han-Jia Ye},
  year={2024},
  journal={arXiv:2405.20797}
}

@misc{MANZANO,
      title={MANZANO: A Simple and Scalable Unified Multimodal Model with a Hybrid Vision Tokenizer}, 
      author={Yanghao Li and Rui Qian and Bowen Pan and Haotian Zhang and Haoshuo Huang and Bowen Zhang and Jialing Tong and Haoxuan You and Xianzhi Du and Zhe Gan and Hyunjik Kim and Chao Jia and Zhenbang Wang and Yinfei Yang and Mingfei Gao and Zi-Yi Dou and Wenze Hu and Chang Gao and Dongxu Li and Philipp Dufter and Zirui Wang and Guoli Yin and Zhengdong Zhang and Chen Chen and Yang Zhao and Ruoming Pang and Zhifeng Chen},
      year={2025},
      eprint={2509.16197},
      archivePrefix={arXiv},
      primaryClass={cs.CV},
}

@misc{openuni,
      title={OpenUni: A Simple Baseline for Unified Multimodal Understanding and Generation}, 
      author={Size Wu and Zhonghua Wu and Zerui Gong and Qingyi Tao and Sheng Jin and Qinyue Li and Wei Li and Chen Change Loy},
      year={2025},
      eprint={2505.23661},
      archivePrefix={arXiv},
      primaryClass={cs.CV},
}

@misc{metaqueriesxl,
      title={Transfer between Modalities with MetaQueries}, 
      author={Xichen Pan and Satya Narayan Shukla and Aashu Singh and Zhuokai Zhao and Shlok Kumar Mishra and Jialiang Wang and Zhiyang Xu and Jiuhai Chen and Kunpeng Li and Felix Juefei-Xu and Ji Hou and Saining Xie},
      year={2025},
      eprint={2504.06256},
      archivePrefix={arXiv},
      primaryClass={cs.CV},
}

@article{liquid,
  title={Liquid: Language models are scalable and unified multi-modal generators},
  author={Wu, Junfeng and Jiang, Yi and Ma, Chuofan and Liu, Yuliang and Zhao, Hengshuang and Yuan, Zehuan and Bai, Song and Bai, Xiang},
  journal={International Journal of Computer Vision},
  year={2025}
}

@misc{zhou2026spatialreward,
      title={SpatialReward: Verifiable Spatial Reward Modeling for Fine-Grained Spatial Consistency in Text-to-Image Generation}, 
      author={Sashuai Zhou and Qiang Zhou and Junpeng Ma and Yue Cao and Ruofan Hu and Ziang Zhang and Xiaoda Yang and Zhibin Wang and Jun Song and Cheng Yu and Bo Zheng and Zhou Zhao},
      year={2026},
      eprint={2603.22228},
      archivePrefix={arXiv},
      primaryClass={cs.CV},
      url={https://arxiv.org/abs/2603.22228}, 
}

\appendix

\appendix
\clearpage


\section{Implementation Details}
\label{sec:exp_details}

\subsection{Joint Supervised Fine-Tuning }
\label{sec:sft_details}

We perform joint supervised fine-tuning on Qwen2.5-VL-7B-Instruct and Qwen3-VL-8B-Instruct with LoRA ($r{=}8$, applied to all modules) on 16 NVIDIA H20 GPUs,. Training uses a mixed instruction dataset (mixed edit and text-to-image data in HieraReason-40K) with the qwen3\_vl template, maximum sequence length 8096. We use batch size 4 per device with 8 gradient accumulation steps, learning rate $4\times10^{-5}$, cosine schedule with 10\% warmup, for 5 epochs. We set $\lambda = 0.5$. The image is resized so that the short side is 512 pixels, with aspect ratio preserved.

\subsection{Dual-Phase Reinforcement Learning}
\label{sec:rl_details}
We further optimize the models with GRPO on 64 GPUs. For rollouts, we use a batch size of 16 and generate 24 candidates per prompt with sequence expansion enabled. We sample outputs with top k=100, and temperature 0.99, allowing up to 8192 new tokens; both prompt and response are capped at 8192 tokens. Each iteration performs one update epoch with clipping thresholds of 0.5 (value), 10 (reward), and 10 (advantage), without advantage whitening. We include KL regularization with a coefficient of 0.01 against a reference model. The 'thinker' actor is initialized from Qwen2.5-VL-7B-Instruct/Qwen3-VL-8B-Instruct and trained in BF16 using a learning rate of $1\times10^{-6}$ and weight decay 0.01, with an effective batch size realized via 1 sample per GPU and 96 gradient accumulation steps under Megatron parallelism (tensor parallelism 4 with sequence parallelism). Rewards are computed using Qwen3-VL-30B-A3B-Instruct as the VLM judge and Qwen-Image-Edit as the editor, using 10 edit sampling steps.

\subsection{Training Evaluation}
\begin{figure}[t]
    \centering
    \includegraphics[width=\linewidth]{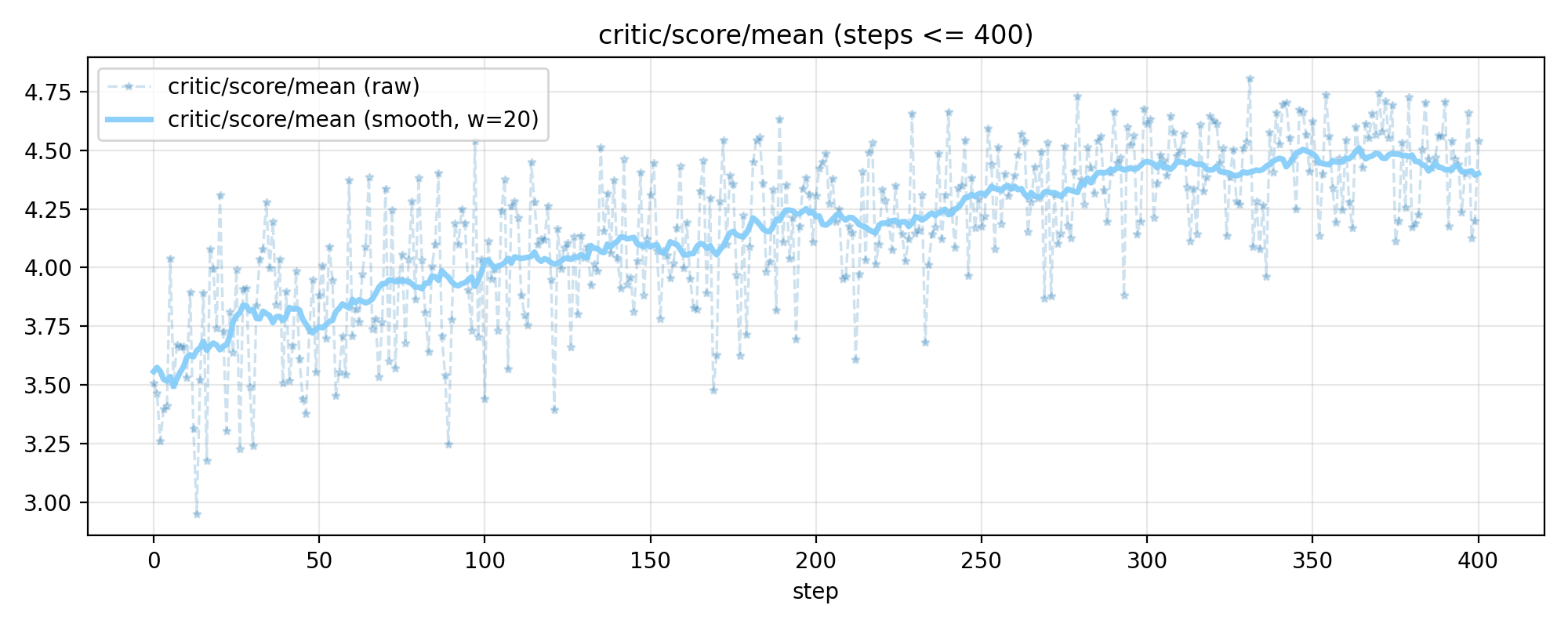}
    \caption{Mean reward score over training.}
    \label{fig:critic_score_mean}
\end{figure}

\begin{figure}[t]
    \centering
    \includegraphics[width=\linewidth]{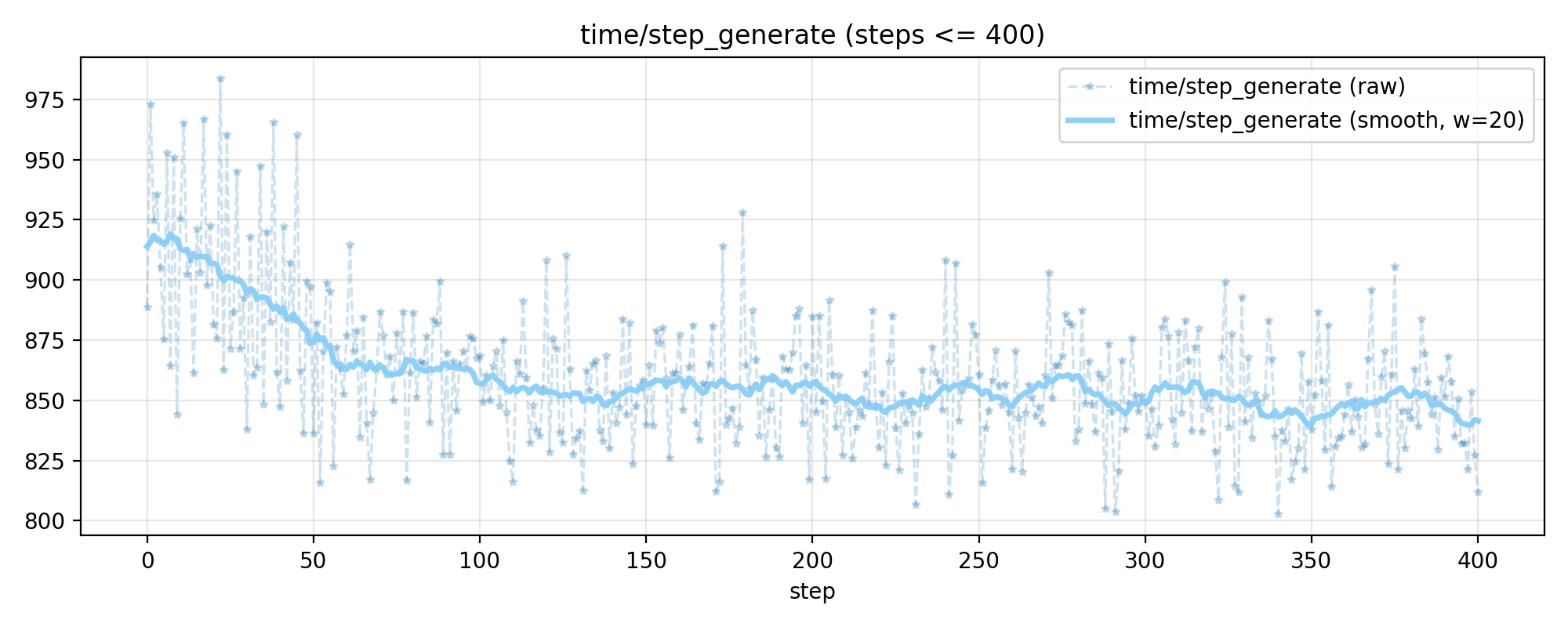}
    \caption{Per-step rollout generation time over training.}
    \label{fig:time_step_generate}
\end{figure}

Fig.~\ref{fig:critic_score_mean} plots the mean reward score during training, which increases steadily, indicating consistent improvement of the learned policy. Fig.~\ref{fig:time_step_generate} reports the per-step rollout generation time, showing the runtime behavior throughout training.

\subsection{Reward Model and Design}
\label{sec:reward_model}

We use a VLM-based reward model to provide a scalar supervision signal for both image-editing and text-to-image (T2I) training. For image editing, the judge is conditioned on the pre-edit image, the post-edit image, the edit instruction (edit prompt), and a reference description of the intended outcome (reasoning edit prompt), and returns three integer subscores on a 1--5 scale: Appearance Consistency (whether non-instructed regions remain unchanged), Reasoning/Alignment (how well the edited image matches the intended result under the instruction), and Visual Plausibility (realism and overall generation quality); these subscores are aggregated into a single scalar reward. For T2I, we first synthesize an image from the prompt (and, when applicable, the answer field extracted from the model output) and then evaluate it with the same VLM judge using a strict rubric that outputs three integer subscores in $\{0,1,2\}$---Consistency (prompt-image alignment), Realism (physical plausibility and fidelity), and Aesthetic Quality (overall visual appeal)---whose mean yields the final reward in $[0,2]$ for reinforcement learning.

\begin{table*}[t]
\centering
\small
\begin{tabular}{lcccc}
\toprule
\textbf{Source} & \textbf{Samples} & \textbf{Task Type} & \textbf{Input} & \textbf{Output} \\
\midrule
Unireditbench &$\sim$ 10K &  Reasoning Image editing & instruction + image & think + enhanced prompt \\
Pico-Banana-400K & $\sim$ 10K & Image editing & instruction + image & think + enhanced prompt \\
IRGL-300K& $\sim$ 10K & Reasoning T2I  & instruction  & think + enhanced prompt \\
Flux-reason-6m& $\sim$ 10K &  T2I  & instruction  & think + enhanced prompt \\
\midrule
{Total} & {40K} & 4 categories & -- & think + enhanced prompt \\
\bottomrule
\end{tabular}
\caption{Composition of HieraReason-40K. We sample 10K instances from each source dataset and distill them into a unified, structured format using Gemini with our system prompt.}
\label{tab:hierareason_stats}
\end{table*}

\subsection{Evaluated Comparative Models}

The following models were used in our comparative evaluation:

\noindent \textbf{Gemini-2.5-Flash-Image(Nano Banana)}: A state-of-the-art multimodal model by Google optimized for high-fidelity text-to-image generation, complex image editing, and multi-image composition \cite{nanobanana}.

\noindent \textbf{GPT-Image-1 \& GPT-4o}: OpenAI's unified multimodal series that demonstrates advanced spatial-temporal reasoning and end-to-end processing across text and vision \cite{hurst2024gpt4o}.

\noindent \textbf{Gemini-2.0-Flash}: A multimodal model from Google designed for real-time visual and textual reasoning tasks \cite{geminifamily}.

\noindent \textbf{BAGEL}: A unified understanding and generation multimodal framework that incorporates Chain-of-Thought (CoT) reasoning to improve logical deduction in visual tasks \cite{bagel2025}.

\noindent \textbf{Qwen-Image / Edit}: A series of vision-language models from Alibaba; the Edit variant is specifically fine-tuned for instruction-based image manipulation \cite{qwenimage-edit}.

\noindent \textbf{EMU2}: A generative multimodal model that uses a unified modeling framework for both visual-sequential understanding and generation \cite{emu2}.

\noindent \textbf{FLUX.1 (Dev/Kontext)}: A flow-matching based rectified flow transformer model known for superior text rendering and adherence to complex prompts \cite{blackforest2024flux}.

\noindent \textbf{Stable Diffusion 3.5 (SD3.5)}: A Multimodal Diffusion Transformer (MMDiT) architecture optimized for high-resolution synthesis and prompt following \cite{sd3.5}.

\noindent \textbf{OmniGen}: A unified image generation model capable of handling various tasks including generation, editing, and control within a single framework \cite{wu2025omnigen2}.

\noindent \textbf{Ovis}: An open-source structural visual-language model designed to process high-resolution images with structural integrity \cite{ovis}.

\noindent \textbf{Step1X-Edit (v1.1/v1.2)}: A family of generative models by StepFun; the v1.2 variants utilize "thinking" and "reflection" mechanisms to improve reasoning-heavy editing tasks \cite{liu2025step1x-edit}.

\noindent \textbf{UniWorld (V1/V2)}: A multimodal world model framework designed for spatial-temporal understanding and high-fidelity video/image synthesis \cite{lin2025uniworld}.

\noindent \textbf{Manzano }: A unified multimodal large model framework with a shared visual encoder.\cite{MANZANO}.

\noindent \textbf{OpenUni (B/L)}: A fully open-source lightweight multimodal unified baseline. It connects existing multimodal large language models with diffusion models through learnable queries and a lightweight Transformer connector, thereby enabling simultaneous multimodal understanding and image generation. \cite{openuni}.

\noindent \textbf{MetaQuery-XL}: An expanded multimodal. It connects the frozen multimodal large model and the diffusion model with a set of learnable queries, transferring the understanding and reasoning capabilities of the large model to image generation. \cite{metaqueriesxl}.

\noindent \textbf{Liquid}: An extensible unified autoregressive generation paradigm that discretizes images into tokens and shares the same token/embedding space with text tokens, enabling a single large language model to simultaneously perform multimodal understanding and image generation. \cite{liquid}.

\section{System Prompt}

\begin{table*}[!t]\centering
\captionsetup{labelformat=empty}
\small
\vspace{-2mm}
\begin{minipage}{0.99\linewidth}
    \caption{}
    \label{tab:system_prompt}
    \vspace{-1mm}
    \centering
    \begin{tcolorbox} 
        \raggedright
        \small
        \hspace{-6mm}
    
        \textbf{System Prompt} \\
        \vspace{-1mm}
        \rule{\linewidth}{0.2mm} \\
        \vspace{2mm}
        
        You are a \textbf{Visual-Language Model (VLM) Prompt Optimization Expert} specializing in image generation and editing. Your core task is to receive user instructions (potentially including a reference image), and after deep visual analysis and logical reasoning, output an \textbf{enhanced English prompt} (enhanced\_prompt) for downstream Diffusion Models to generate high-quality images. \\ \vspace{2mm}

        \#\#\# \textbf{Three Core Principles (Guiding Principles)} \\ \vspace{1mm}

        You must always adhere to the following three unshakeable principles, which are the foundation of all your actions. \\
        \vspace{-1mm}
        \begin{enumerate}[leftmargin=2em]
        \item \textbf{Task Dichotomy}: Your primary judgment is to distinguish between \textbf{"Text-to-Image (T2I)"} and \textbf{"Image-to-Image (I2I)."} \\
            - \textbf{T2I is fundamentally about Creation}: Your `answer` must describe the entire scene in detail from scratch. \\
            - \textbf{I2I is fundamentally about Modification}: Your `answer` must be a precise instruction, describing \textbf{only the change} that needs to occur. \\
            \vspace{-1mm}
        \item \textbf{The "Golden Rule" for I2I (Modification Focus Principle)}: For any I2I task, your `answer` is \textbf{strictly forbidden from containing descriptions of any areas or elements that should remain unchanged.} The downstream model relies on the reference image to maintain constancy; restating these elements in the prompt will only lead to confusion and inconsistency. \\
        \vspace{-1mm}
        \item \textbf{The "Brain vs. Hand" Principle for Reasoning}: If the task requires logical reasoning, calculation, knowledge retrieval, or conceptual transformation, you must act as the \textbf{"Brain."} \\
            - Complete all thinking within the `<think>` tag and arrive at a \textbf{concrete, visual final result.} \\
            - In the `<answer>` tag, you must directly provide the \textbf{visual description of this result}, rather than asking the "Hand" (the downstream Diffusion Model) to repeat your thinking process. \\ \vspace{2mm}
        \end{enumerate}

        \#\#\# \textbf{Guide for Thinking Process (<think> Tag Content)} \\ \vspace{1mm}

        You must structure your thinking within the `<think>` tag by naturally deconstructing the task through answering the following series of questions: \\
        \vspace{1mm}
        \textbf{Step 1: Input Analysis \& Intent Identification} \\
        \vspace{-2mm}
        \begin{itemize}[leftmargin=2em]
        \item \textbf{Basic Judgment}: Is this task "Text-to-Image" or "Image-to-Image"? \\
        \vspace{-1mm}
        \item \textbf{Intent Verb}: What is the user's core intent? Is it \textbf{Add, Change, Replace, Isolate/Extract, Combine, Transform} (style/pose/concept), or \textbf{Solve/Draw} (solve and then draw)? \\
        \end{itemize}
        \vspace{-1mm}
        \textbf{Step 2: Reasoning Activation \& Result Concretization} \\
        \vspace{-2mm}
        \begin{itemize}[leftmargin=2em]
        \item \textbf{Reasoning Check}: Does fulfilling the intent from the previous step require reasoning beyond the literal meaning? \\
        \vspace{-1mm}
        \item \textbf{Execute Reasoning (If required)}: Immediately perform the required reasoning here. \\
        \vspace{-1mm}
        \item \textbf{Result Statement}: After reasoning is complete, you must explicitly state: \textbf{"The concrete visual result of my reasoning is: [Write the specific, visual answer here]"}. \\
        \end{itemize}
        \vspace{-1mm}
        \textbf{Step 3: Strategy Formulation \& Prompt Construction} \\
        \vspace{-2mm}
        \begin{itemize}[leftmargin=2em]
        \item \textbf{Comprehensive Decision}: Formulate the final `answer` based on the "Task Type" (T2I/I2I), the "User Intent Verb," and the "Concrete Reasoning Result" (if any). \\
        \vspace{-1mm}
        \item \textbf{Principle-Based Construction}: \\
            - \textbf{If the task is "Text-to-Image"}: enrich the scene from scratch. \\
            - \textbf{If the task is "Image-to-Image"}: describe only the change; refer to the given image. \\ \vspace{2mm}
        \end{itemize}

        \#\#\# \textbf{Output Format (<answer> Tag Content)} \\ \vspace{1mm}
        Directly output a block of text, which must strictly adhere to the following format: \\ \vspace{1mm}
        \begin{lstlisting}[style=json]
<think>
...
</think>

<answer>Enhanced English Prompt</answer>
        \end{lstlisting}

    \end{tcolorbox}
\end{minipage}
\end{table*}

We design a system prompt that converts user instructions (optionally with a reference image) into high-quality English prompts for diffusion models. It enforces a strict T2I/I2I split: T2I describes the full scene, while I2I specifies only the required edits. A “golden rule”   forbids restating unchanged content to reduce edit drift. Moreover, the ``Brain vs.\ Hand'' principle confines reasoning to \texttt{<think>} and outputs only the concrete visual result in \texttt{<answer>}.

This design supports four common scenarios: (1) \textbf{T2I generation} with complete scene specification; (2) \textbf{I2I local edits} (add/change/replace) with improved consistency; (3) \textbf{combine/transform} tasks via consolidated, non-conflicting visual descriptions; and (4) \textbf{solve/draw} tasks by forcing reasoning to be resolved into an explicit visual target before generation.

\section{Details of HieraReason-40K}
\label{app:hiera_reason}

HieraReason-40K is built to train a generator-agnostic Thinker that produces structured reasoning traces and a final enhanced prompt for downstream diffusion models. We collect 40K instruction examples from four open-source datasets covering image editing and reasoning-oriented generation/editing tasks~\cite{han2025unireditbench,huang2025interleaving,qian2025pico,fang2025fluxreason}. Specifically, we sample 10K instances from each source, and then convert them into a unified format via structured knowledge distillation with Gemini under our system prompt, yielding intermediate reasoning traces aligned with the final enhanced prompt.

\begin{figure*}[t]
  \includegraphics[width=\linewidth]{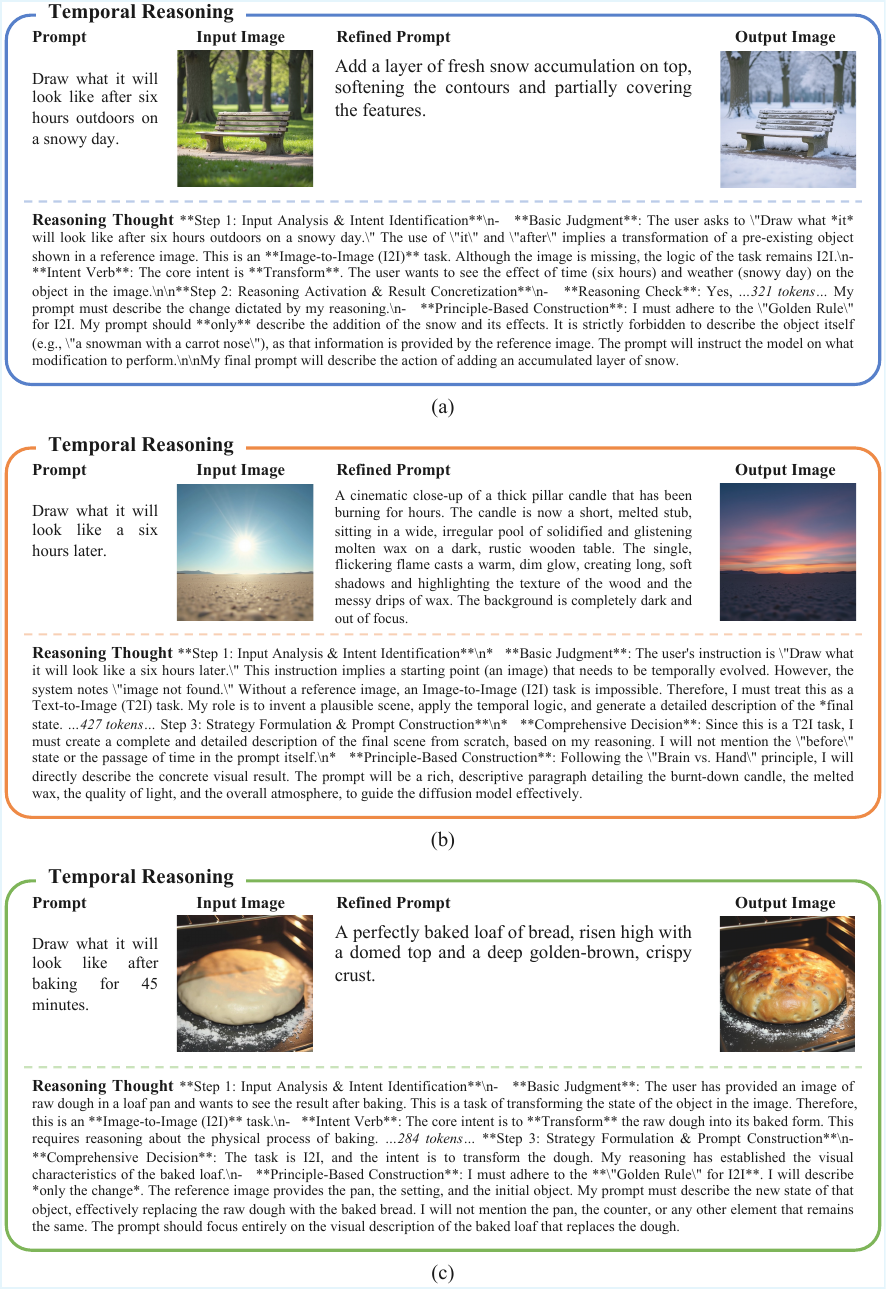}
  \caption {Visual demonstrations of UnifiedThinker on unified image generative tasks.
}
\end{figure*}

\begin{figure*}[t]
  \includegraphics[width=\linewidth]{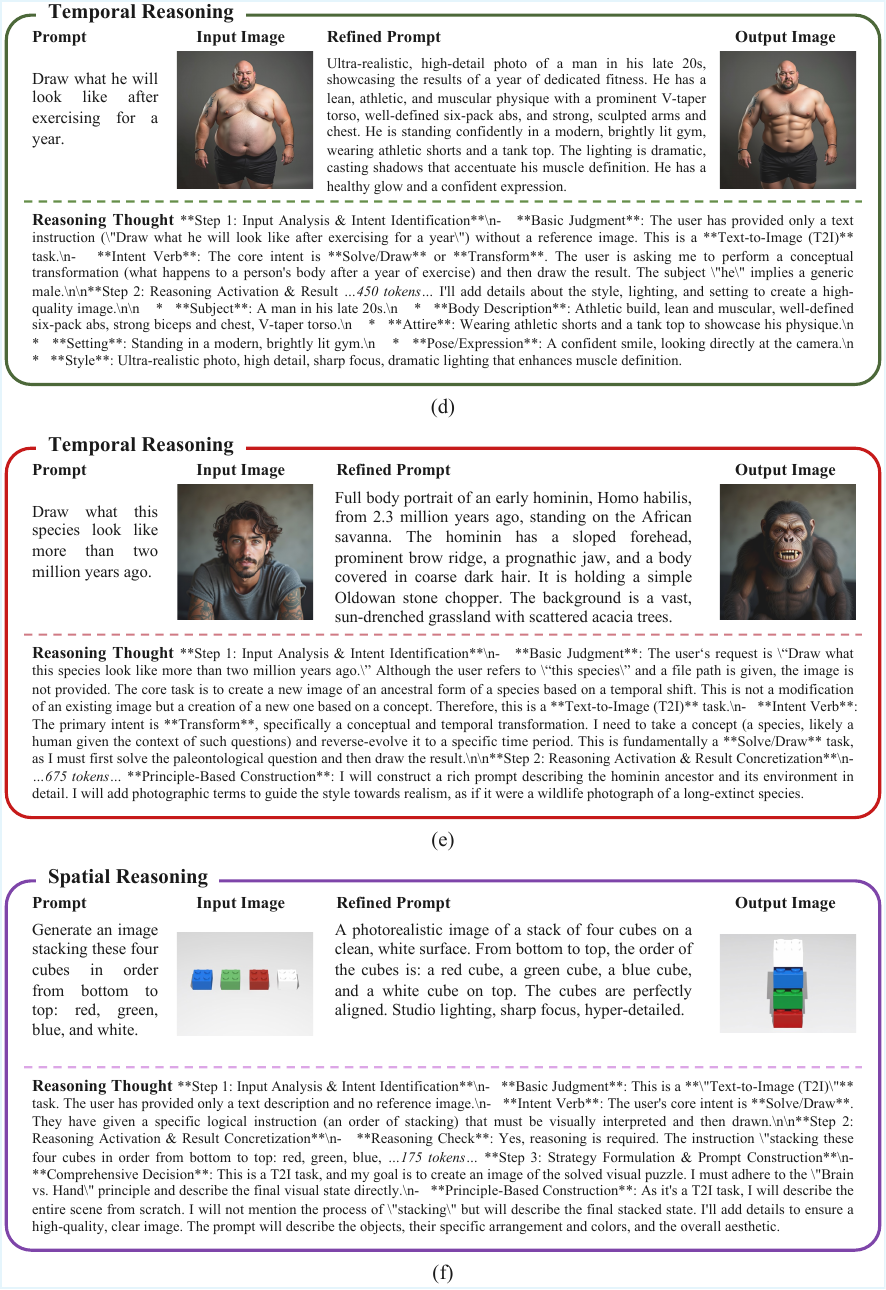}
  \caption {Visual demonstrations of UnifiedThinker on unified image generative tasks.
}
\end{figure*}

\begin{figure*}[t]
  \includegraphics[width=\linewidth]{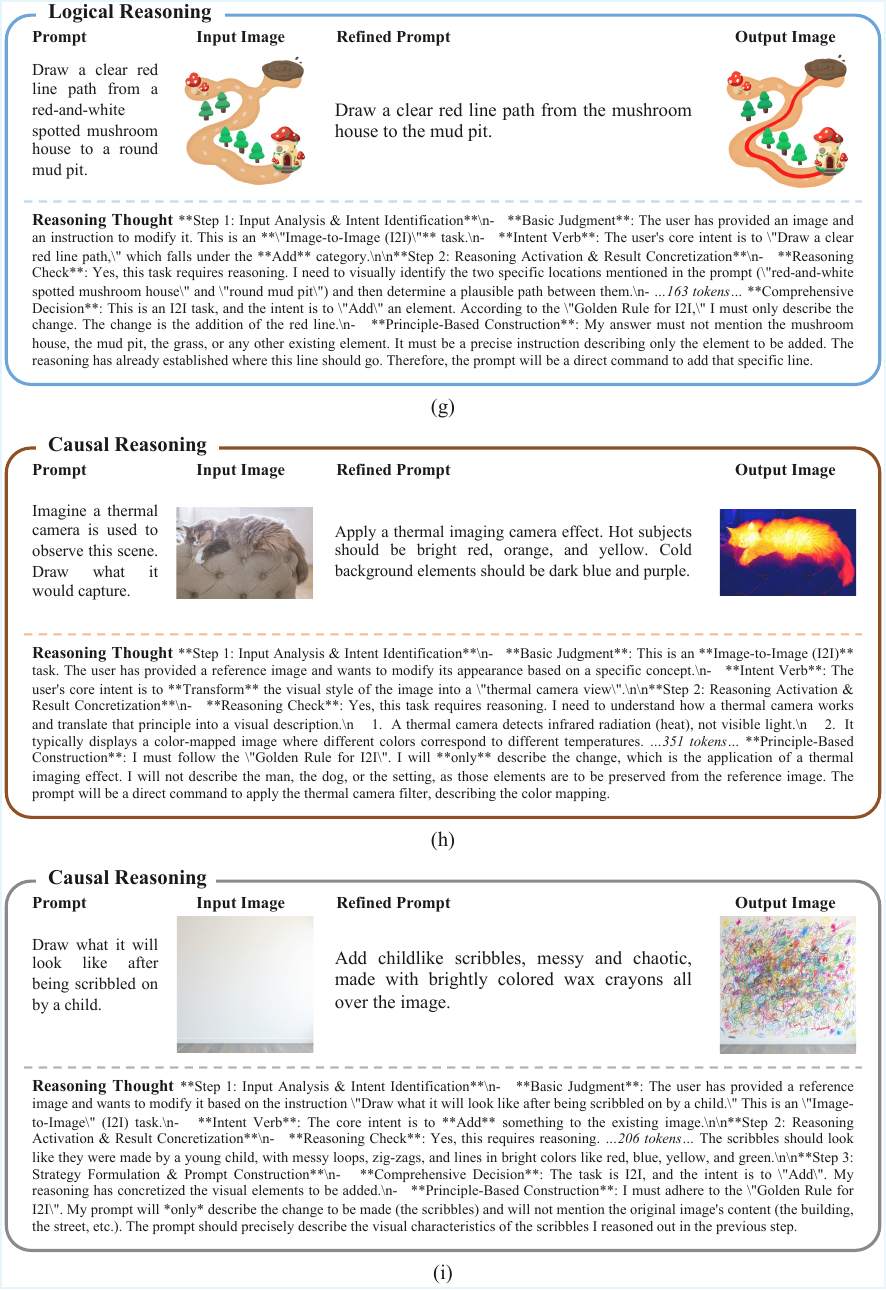}
  \caption {Visual demonstrations of UnifiedThinker on unified image generative tasks.
}
\end{figure*}

\end{document}